\documentclass{article} % For LaTeX2e
\usepackage{iclr2026_conference,times}

\usepackage{amsmath,amsfonts,bm}

\def\eqref#1{equation~\ref{#1}}
\def\1{\bm{1}}

\DeclareMathAlphabet{\mathsfit}{\encodingdefault}{\sfdefault}{m}{sl}
\SetMathAlphabet{\mathsfit}{bold}{\encodingdefault}{\sfdefault}{bx}{n}

\usepackage{hyperref}
\usepackage{url}
\usepackage{graphicx}
\usepackage{amssymb}
\usepackage{amsmath}
\usepackage{booktabs}
\usepackage{bm}
\usepackage{xcolor}
\usepackage{multirow}
\usepackage{algorithm}
\usepackage{algpseudocode}
\usepackage{amsfonts}
\usepackage{colortbl}
\usepackage{makecell}
\usepackage{bbding}
\usepackage{caption}
\usepackage{subcaption}

\title{Learning Domain-Aware Task Prompt Representations for Multi-Domain All-in-One Image Restoration}

\iclrfinalcopy

\author{%
Guanglu Dong$^{1,}$\thanks{Work done during an internship at MedAI Technology (Wuxi) Co. Ltd.} \quad
Chunlei Li$^{2}$ \quad
Chao Ren$^{1}$ \quad
Jingliang Hu$^{2}$ \quad
Yilei Shi$^{2}$ \\[4pt]
\textbf{Xiao Xiang Zhu}$^{3}$ \quad
\textbf{Lichao Mou}$^{2,}$\thanks{Corresponding author.} \\[4pt]
{\small
$^{1}$Sichuan University \quad
$^{2}$MedAI Technology (Wuxi) Co. Ltd. \quad
$^{3}$Technical University of Munich
}\\[4pt]
{\small
\texttt{dongguanglu@stu.scu.edu.cn} \quad
\texttt{lichao.mou@medimagingai.com}
}
}

\begin{document}

\maketitle
\begin{abstract}

Recently, significant breakthroughs have been made in all-in-one image restoration (AiOIR), which can handle multiple restoration tasks with a single model. However, existing methods typically focus on a specific image domain, such as natural scene, medical imaging, or remote sensing. In this work, we aim to extend AiOIR to multiple domains and propose the first multi-domain all-in-one image restoration method, DATPRL-IR, based on our proposed \textit{\textbf{D}omain-\textbf{A}ware \textbf{T}ask \textbf{P}rompt \textbf{R}epresentation \textbf{L}earning}. Specifically, we first construct a task prompt pool containing multiple task prompts, in which task-related knowledge is implicitly encoded. For each input image, the model adaptively selects the most relevant task prompts and composes them into an instance-level task representation via a prompt composition mechanism (PCM). Furthermore, to endow the model with domain awareness, we introduce another domain prompt pool and distill domain priors from multimodal large language models into the domain prompts. PCM is utilized to combine the adaptively selected domain prompts into a domain representation for each input image. Finally, the two representations are fused to form a domain-aware task prompt representation which can make full use of both specific and shared knowledge across tasks and domains to guide the subsequent restoration process. Extensive experiments demonstrate that our DATPRL-IR significantly outperforms existing SOTA image restoration methods, while exhibiting strong generalization capabilities. Code is available at \url{https://github.com/GuangluDong0728/DATPRL-IR}.
\end{abstract}

\section{Introduction}
\label{introduction}

\begin{figure}[h]
  \centering
  \setlength{\abovecaptionskip}{0.2cm}
  \includegraphics[width=0.86\textwidth]{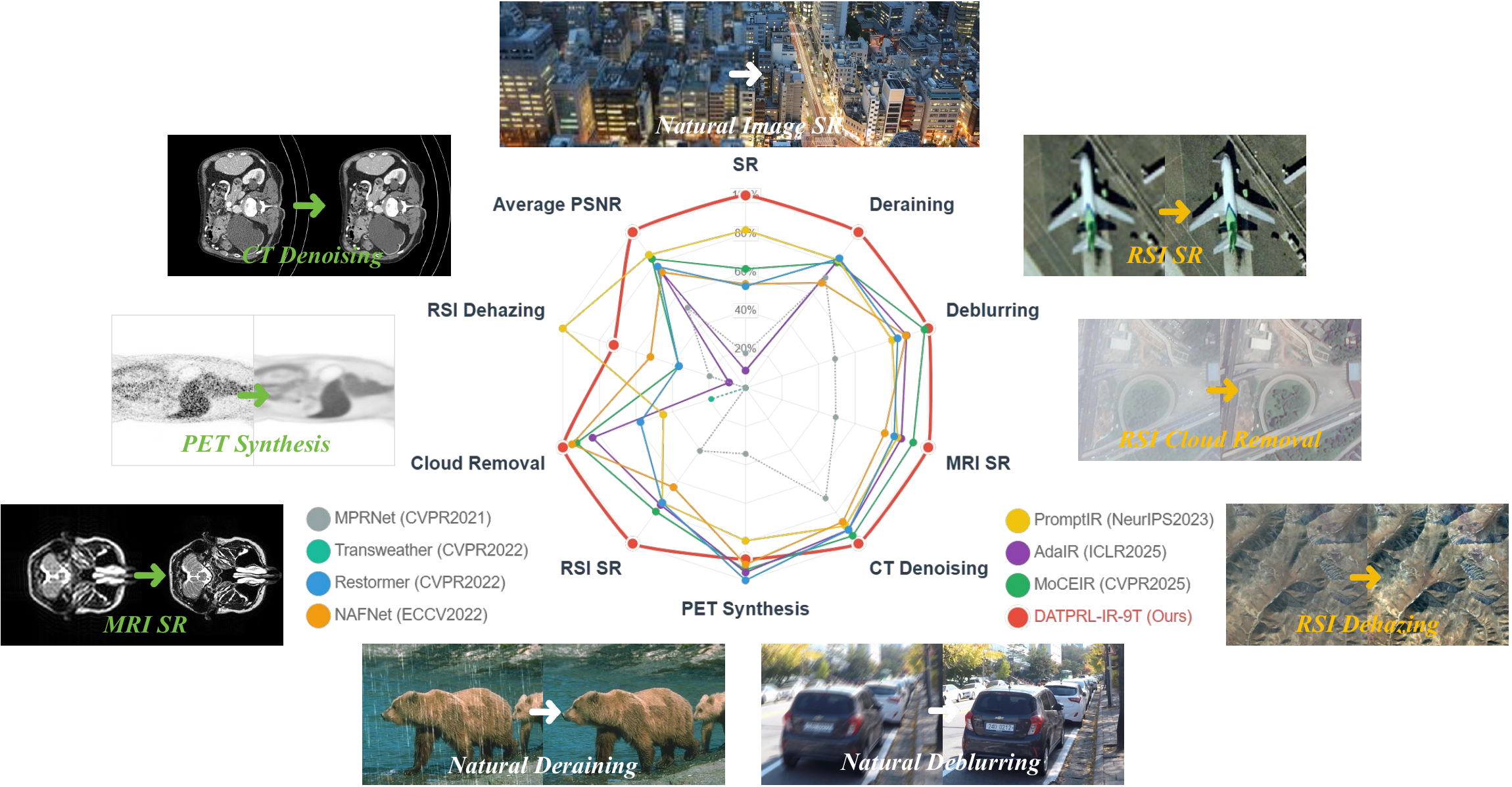}
  \caption{This paper makes a preliminary exploration of multi-domain all-in-one image restoration (MD-AiOIR), aiming at further extending the restoration capability of a single model to a broader range of tasks and image domains, including natural scene, medical imaging, and remote sensing.}
  \label{fig.1}
\end{figure}

Image restoration~\citep{mambair, NAFNet, restormer, SWINIR, MPRNet} has long been a fundamental research in computer vision, aiming to recover high-quality images from their degraded versions. With the advancement of deep learning, image restoration has found widespread applications across multiple image domains, including natural scene, medical imaging, and remote sensing. Early explorations mainly focused on designing independent models for different tasks within each domain, such as natural image super-resolution (SR)~\citep{SRCNN}, natural image deraining~\citep{CSUD}, natural image deblurring~\citep{deblurring1}, CT denoising~\citep{MedIR_CT_denoising}, MRI SR~\citep{MedIR_MRI_SR}, PET synthesis~\citep{MedIR_PET_synthesis}, remote sensing image (RSI) SR~\citep{remote_sensing_IR1}, RSI cloud removal~\citep{remote_sensing_IR3}, RSI dehazing~\citep{remote_sensing_dehazing}, etc.. However, training separate models is undoubtedly time-consuming and resource-intensive, and this greatly limits their applicability in complex real-world scenarios. 

To address the above challenge, all-in-one image restoration (AiOIR)~\citep{airnet, adair, promptir, moceir, instructir, Perceive-IR} has gained increasing attention in recent years, as it seeks to provide a unified solution for handling multiple restoration tasks with a single model. AiOIR first emerge in natural scene, by leveraging implicit or explicit prompts~\citep{promptir, instructir}, contrastive learning~\citep{airnet}, degradation classification~\citep{DCPT}, prior information~\citep{prior_allinone}, or mixture-of-experts (MoE) architecture~\citep{moceir} to enable the restoration networks to better distinguish between different tasks. Meanwhile, inspired by the progress in natural scene, AiOIR methods have also gained popularity in medical imaging~\citep{all-in-one-medir2025, all-in-one-medir-Restore-rwkv, amir}. Though existing methods have achieved remarkable success, they primarily focus on maximizing task differences within a single domain and often overlook the shared commonalities across tasks. They also do not consider the differences or connections between different image domains. When faced with more restoration tasks and image domains, they will face increased learning difficulty. 

In this work, we make the first exploration of multi-domain all-in-one image restoration (MD-AiOIR), aiming to unify diverse restoration tasks across multiple domains within a single model. Inspired by the concept of prompt pool in L2P~\citep{L2P}, we propose domain-aware task prompt representation learning (DATPRL), which adopts a dual-prompt-pool design to learn prompt representations that carry both task-relevant and domain-relevant knowledge. Based on DATPRL, we introduce the first MD-AiOIR method, DATPRL-IR. Specifically, we first construct a task prompt pool with numerous task prompts. For each input image, our DATPRL-IR can adaptively select the most relevant task prompts through a similarity based query mechanism. To express more diverse instance-level information, we propose a prompt composition mechanism (PCM) to combine the selected task prompts into a task prompt representation. The task prompts are optimized jointly with the restoration objectives, ensuring the learning of task-specific knowledge while allowing knowledge sharing across tasks. Additionally, to endow the model with domain awareness, we build a separate domain prompt pool to store domain-related knowledge. We leverage the powerful image understanding ability of multimodal large language models (MLLMs) and employ a cross‑modal alignment to distill domain priors from MLLMs (e.g., LLaVA~\citep{LLaVa1.5}) into the domain prompts. Similarly, our DATPRL-IR will adaptively select the most relevant domain prompts for each input image, and then apply PCM to combine them into an instance-level domain prompt representation. The two prompt representations are then fused into the final domain-aware task representation to guide the subsequent restoration process. Our method effectively exploits the shared knowledge across different tasks and domains, significantly reducing the learning difficulty and facilitating performance improvement across tasks. As illustrated in Figure \ref{fig.1}, under the guidance of domain-aware task prompt representations, our DATPRL-IR significantly surpasses existing methods, demonstrating strong generalization capability. 

Our main contributions can be summarized as follows: (1) To the best of our knowledge, we propose the first multi-domain all-in-one image restoration method, DATPRL-IR, which can handle diverse restoration tasks across multiple domains. (2) Through the proposed domain-aware task prompt representation learning, our DATPRL-IR effectively leverages both specific and shared knowledge across tasks and domains to guide the restoration process. (3) Extensive experiments demonstrate that our method outperforms existing SOTA image restoration approaches on MD-AiOIR tasks. 

\section{RELATED WORK}
\label{related_work}

\textbf{Single-Task Image Restoration.} With the development of deep learning~\citep{CNN}, image restoration techniques have made continuous progress across multiple imaging domains, including natural scene~\citep{SRCNN, yunjin_chen_denoising, CSUD, ntire2025}, medical imaging~\citep{MedIR_MRI_SR, MedIR_CT_denoising, MedIR_PET_synthesis}, and remote sensing~\citep{remote_sensing_IR1, remote_sensing_IR2, remote_sensing_IR3}. By leveraging specific designs for different domains and tasks, a wide variety of restoration sub-tasks have flourished. Recently, with the growing demand for multi-task image restoration and the continuous evolution of foundation backbones (e.g., CNNs~\citep{resnet, UNet}, Transformers~\citep{Transformer, ViT}, and Mamba~\citep{Mamba, Vision_mamba}), a series of general image restoration baselines~\citep{MPRNet, SWINIR, restormer, Uformer, NAFNet, mambair, MaIR} have also emerged, which are capable of handling diverse types of degradations within a unified model architecture. However, these methods require training a separate model for each individual task, which is time-consuming and labor-intensive.

\textbf{All-in-One Image Restoration.} To overcome the limitations above, various all-in-one image restoration (AiOIR) frameworks~\citep{airnet, adair, instructir, promptir, moceir} have continuously emerged and achieved sustained breakthroughs, especially in the natural image domain. AirNet~\citep{airnet} is the first to achieve AiOIR through contrastive learning~\citep{contrastive_learning, MOCO}. IDR~\citep{IDR} integrates degradation-specific priors into the restoration process to enhance performance. PromptIR~\citep{promptir} uses learnable prompt components to encode different degradation information. DA-CLIP~\citep{DA-CLIP} decouples degradation and content semantics based on CLIP~\citep{CLIP}, making the model more sensitive to various degradation knowledge. InstructIR~\citep{instructir} guides the image restoration model through human-written instructions. MoCE-IR~\citep{moceir} introduces complexity experts within a mixture-of-experts (MoE) architecture to efficiently allocate task-specific resources. DCPT~\citep{DCPT} propose a degradation classification pre-training strategy to classify the degradation type of input images. In addition to the methods for natural domain, recent AiOIR techniques have also started to gain attention in the field of medical imaging~\citep{amir, all-in-one-medir-Restore-rwkv, all-in-one-medir2025}. However, current research mainly focuses on exploring a specific domain, with most approaches aiming to better distinguish between different tasks while overlooking the commonalities between them. 

\textbf{Prompt Learning-based Image Restoration.} Inspired by the success of prompt learning in natural language processing~\citep{prompt_nlp1, prompt_nlp2}, high-level computer vision~\citep{L2P, dualprompt}, and multi-modal models~\citep{prompt_mutimodal1, prompt_mutimodal2}, it has also been widely applied in image restoration recently. PromptRestorer~\citep{Promptrestorer} takes the advantage of prompt learning to perceive degradation, achieving progress on individual tasks such as image deraining, deblurring, and dehazing. SFD~\citep{SFD} trains learnable antonymous prompt pairs in an adversarial manner to promote global discrimination for super-resolution images. PromptIR~\citep{promptir} is the first to explore the capability of prompt learning in all-in-one image restoration, and it subsequently inspires a series of prompt-based all-in-one restoration methods~\citep{prompt-all-in-one-tcsvt, all-in-one-Prompt, FrePrompter, Prores, MPerceiver, instructir}, which employ explicit or implicit prompts to guide restoration process. In this work, different from existing methods, we propose a novel domain-aware task prompt representation learning method, which effectively leverages both both the specific and shared knowledge across restoration tasks and image domains to guide multi-domain all-in-one image restoration.
\vspace{-0.05cm}
\section{DATPRL-IR for MD-AiOIR}
\label{method}

\begin{figure}[t]
  \centering
  \setlength{\abovecaptionskip}{0.2cm}
  \includegraphics[width=1.0\textwidth]{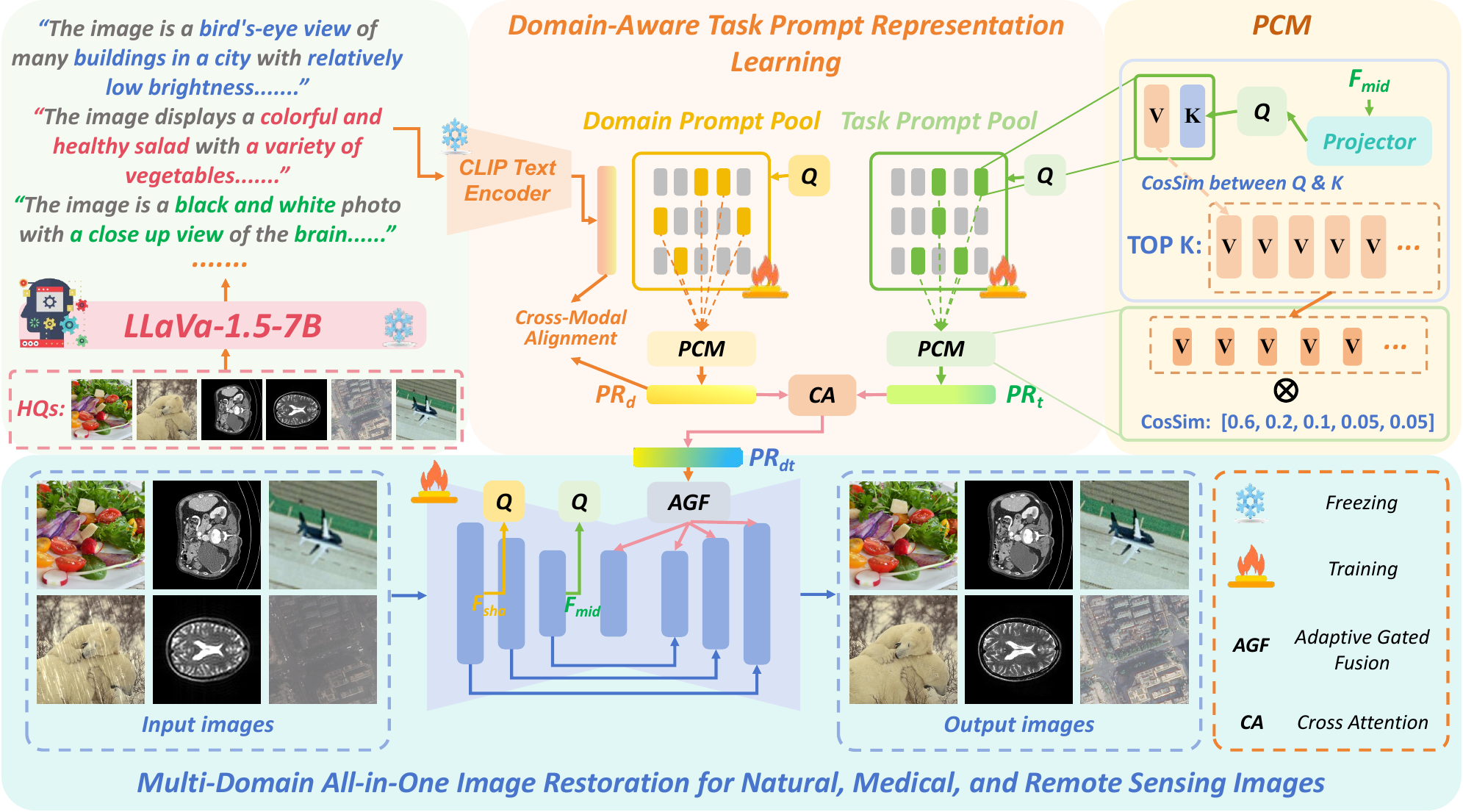}
  \caption{Framework of the proposed DATPRL-IR for multi-domain all-in-one image restoration. By introducing domain-aware task representation learning, DATPRL-IR can fully utilize both specific and shared knowledge across tasks and domains, effectively reducing the learning difficulty of the model and improving its performance.}
  \label{fig.framework}
\end{figure}

\subsection{Domain-Aware Task Prompt Representation Learning}
\label{Domain-Aware Task Prompt Representation Learning}

\textbf{Motivation.}
In this work, we aim to take the first step towards Multi-Domain All-in-One Image Restoration (MD-AiOIR), extending AiOIR to more restoration tasks across multiple image domains. A key challenge is how to alleviate the learning difficulties introduced by the increasing number of restoration tasks and image domains. Prior studies~\citep{AdaIR2, IDR} have found that different image restoration tasks share certain inherent commonalities or similar latent representations and there is a certain mutual promotion effect between different restoration tasks~\citep{instructir}, such as super-resolution and motion deblurring. Additionally, though images from different domains exhibit their own distinct visual characteristics, they share some common terms. Combining these specific and shared visual characteristics can facilitate the discrimination of the image domain. For instance, “grayscale + human organs” typically corresponds to medical images, while “bird’s-eye view + buildings” often indicates remote sensing scenarios. Therefore, we infer that effectively leveraging both the specific and shared knowledge across tasks and domains can help reduce the learning difficulty and further enhance the restoration performance. Inspired by L2P~\citep{L2P}, prompt pools offer an effective way to encode and organize both specific and shared knowledge. Building on this insight, we propose \textit{\textbf{D}omain-\textbf{A}ware \textbf{T}ask \textbf{P}rompt \textbf{R}epresentation \textbf{L}earning} and introduce the first MD-AiOIR method, DATPRL-IR. 

\textbf{Overall Framework.} As illustrated in Figure \ref{fig.framework}, our DATPRL-IR mainly consists of an encoder–decoder based restoration backbone, a task prompt pool, a domain prompt pool, a CLIP~\citep{CLIP} text encoder, and the LLaVA-1.5-7B~\citep{LLaVa1.5} model. The task and domain prompt pools store $N_{t}$ and $N_{d}$ prompts, respectively. These prompts implicitly encapsulate knowledge related to restoration tasks and image domains. Given a degraded input image, our model adopts a query–retrieval–composition paradigm to adaptively query both prompt pools to retrieve the most relevant prompts for the task and domain, which are then composed into two representations: a task prompt representation $\mathbf{PR}_{\text{t}}$ and a domain prompt representation $\mathbf{PR}_{\text{d}}$. Subsequently, these two representations are integrated through a cross-attention mechanism to produce a domain-aware task prompt representation $\mathbf{PR}_{\text{dt}}$, which can effectively guide the restoration process.

\textbf{Task Prompt Pool.} Task prompt (TP) pool is used to implicitly store both specific and shared knowledge across different tasks, and each prompt in TP pool is represented as a pair of a key $\mathbf{K}_j^{\text{task}} \in \mathbb{R}^d$ and value $\mathbf{V}_j^{\text{task}} \in \mathbb{R}^{T \times d}$. We use a learnable projector to map the middle feature $F_{mid}$ of the input image extracted by the encoder of the restoration network into a query $\mathbf{Q}^{\text{task}} \in \mathbb{R}^d$ with the same dimension as $\mathbf{K}_j^{\text{task}}$. Based on the cosine similarities $s_j^{\text{task}}$ between the $\mathbf{Q}^{\text{task}}$ and each $\mathbf{K}_j^{\text{task}}$, the top k most relevant values $\mathbf{V}_k^{\text{task}}$ can be retrieved from the TP pool. To enable the limited set of prompts to provide more diverse instance-level prompt guidance, we design a prompt composition mechanism (PCM) to combine the selected $\mathbf{V}_k^{\text{task}}$ into an instance-level task prompt representation $\mathbf{PR}_{\text{t}}$ according to the similarity scores $s_k^{\text{task}}$:
\begin{equation}
\alpha_j^{\text{task}} = \frac{\exp(s_j^{\text{task}}/T_{\text{task}})}{\sum\limits_{\ell \in k} \exp(s_\ell^{\text{task}}/T_{\text{task}})}, \  \ \mathbf{PR}_{\text{t}} = \sum_{j \in k} \alpha_j^{\text{task}} \mathbf{V}_{j}^{\text{task}}.
\end{equation}
where $\alpha_j^{\text{task}}$ denotes the relative cosine similarity among the selected prompts, and $T_{\text{task}}$ is the temperature parameter. During training, task prompts are optimized jointly with restoration objectives, ensuring the learning of task-related knowledge while allowing knowledge sharing across tasks.

\begin{figure}[t]
  \centering
  \setlength{\abovecaptionskip}{0.2cm}
  \includegraphics[width=0.98\textwidth]{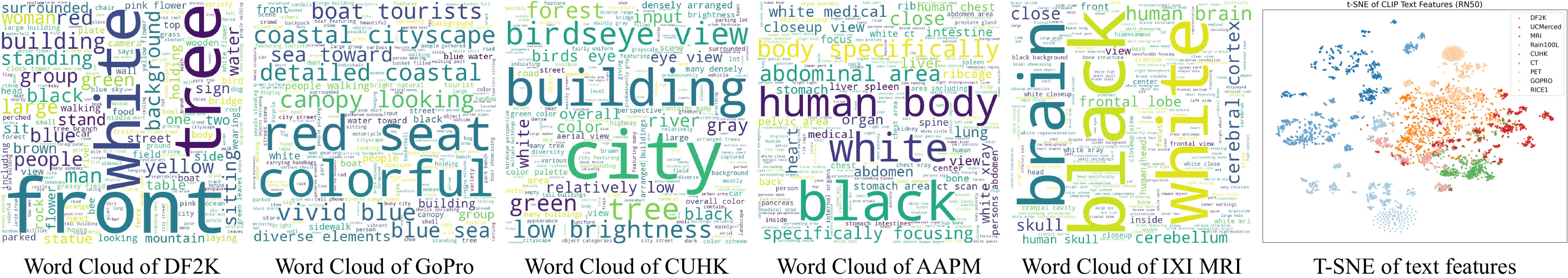}
  \caption{A partial visualization of the word clouds generated from the text descriptions produced by LLAVA, and the t-SNE clustering analysis of the text descriptions corresponding to the 9 datasets from different domains and tasks. It can be observed that images from different domains exhibit their own characteristics while also sharing certain overlapping features.}
  \label{fig.Wordcloud}
\end{figure}

\textbf{Domain Prompt Pool.} As shown in Figure \ref{fig.Wordcloud}, images from different domains exhibit their own characteristics while also sharing certain overlapping features. Integrating these characteristics will help the model to better determine the domain of an input image and to enrich its domain knowledge. To this end, we construct another domain prompt (DP) pool to store and organize such domain priors. The domain prompts in the DP Pool are also constructed as a set of key–value pairs. For each input image, we use another projector to map the shallow features extracted from the first layer of the restoration backbone into a domain query $\mathbf{Q}^{\text{dom}}$. Based on the cosine similarity between $\mathbf{Q}^{\text{dom}}$ and the key $\mathbf{K}_j^{\text{dom}}$ of each domain prompt, we select the top k most relevant prompts. Similar to the TP pool, we also employ the PCM to combine the value $\mathbf{V}_j^{\text{dom}}$ of these candidates into an instance-level domain prompt representation $\mathbf{PR}_{\text{d}}$. To endow $\mathbf{PR}_{\text{d}}$ with rich and interpretable domain knowledge, we employ LLaVA‑1.5‑7B to generate multi-perspective descriptions for the high-quality images (HQs) corresponding to each input, covering aspects such as image content, color richness, object category, brightness, and camera/viewpoint. During training, these textual descriptions are fed into the CLIP text encoder to obtain the corresponding text features $\mathbf{F}_{\text{text}}$, and the training of the DP Pool is constrained by the following cross-modal alignment loss:

\begin{equation}
\mathcal{L}_{\text{align}} = \frac{1}{B}\sum_{n=1}^{B} \Big(1 - \cos(\mathbf{PR}_{\text{dom}}^n, \mathbf{F}_{\text{text}}^n\Big).
\end{equation}
where $B$ denotes the batch size. During the joint training of our model, $\mathcal{L}_{\text{align}}$ encourages the domain prompts to capture both specific and share domain knowledge that benefits the restoration objective, providing domain-related information to the network. It is worth noting that the LLaVA and CLIP will not be used during the inference stage without introducing any additional inference overhead. 

\textbf{Domain-Aware Task Prompt Representation.} Finally, the task and domain prompt representations $\mathbf{PR}_{\text{t}}$ and $\mathbf{PR}_{\text{d}}$ will be fused through a cross-attention layer to learn a domain-aware task prompt representation $\mathbf{PR}^{\text{dt}}$, which will be used to guide the restoration process. Considering that different layers of the restoration backbone may have varying demands for prompt information, inspired by UniECS~\citep{UniECS}, we dynamically control the contribution ratio between backbone features and $\mathbf{PR}_{\text{dt}}$ at each layer through an adaptive gated fusion (AGF):
\begin{equation}
\mathbf{F}^e_l 
= \mathrm{CrossAttn}\!\left(
    \alpha_l \, \mathbf{F}_l, \;
    (1 - \alpha_l) \, \mathbf{PR}_{\text{dt}}
\right),
\label{eq:AGF}
\end{equation}

where $\mathbf{F}^e_l$ and $\mathbf{F}_l$ denote the enhanced feature map and the pre-fusion feature map at the $l$-th layer respectively, and $\alpha_l \in [0,1]$ is a learnable gating coefficient for the $l$-th layer. AGF allows each layer to independently learn the optimal fusion ratio, enabling a adaptive integration of $\mathbf{PR}^{\text{dt}}$ and $\mathbf{F}_l$.

\subsection{Prompt Pool Regularization}
\label{Prompt Pool Regularization}

We introduce a series of regularization terms to avoid the model degenerating into undesirable behaviors: e.g., over-relying on a small subset of prompts, or learning redundant or highly correlated prompt contents, especially during the early training phase. Firstly, we adopt a diversity regularization to encourage diversity of the learned prompts. Given a prompt pool with $N$ prompts, we first compute the pairwise cosine similarity matrix of their values $\mathbf{V} \in \mathbb{R}^{T \times d}$:
\begin{equation}
\mathbf{S}_{ij} = \frac{\mathbf{V}_i \cdot \mathbf{V}_j^\mathbf{T}}{\|\mathbf{V}_i\|_2 \, \|\mathbf{V}_j\|_2},
\end{equation}
where $\mathbf{S}_{ij}$ is the pairwise cosine similarity. To exclude self-similarity, we apply a mask $\mathbf{M} = \mathbf{I} - \mathbf{I}_N$, where $\mathbf{I}_N$ is the $N \times N$ identity matrix. The diversity regularization loss is then formulated as:
\begin{equation} 
\mathcal{L}_{\mathrm{div}} = \frac{1}{N(N-1)} \sum_{i=1}^{N} \sum_{j=1}^{N} \mathbf{M}_{ij} \cdot \max\big(0, \mathbf{S}_{ij} - \tau_{\mathrm{div}}\big),
\end{equation}
where $\tau_{\mathrm{div}}$ is a predefined similarity threshold. Minimizing $\mathcal{L}_{\mathrm{div}}$ encourages prompts to occupy distinct regions in the representation space, avoiding collapse to similar contents. 

Furthermore, we adopt a prompt entropy regularization to encourage more balanced utilization across the prompts. Given a query $\mathbf{q} \in \mathbb{R}^d$ mapped from the input image and a pool of $P$ prompt keys $\{\mathbf{k}_j\}_{j=1}^P$, we first compute the selection probability $p_j$ of each prompt based on the cosine similarity score $s_j$ between $\mathbf{q}$ and $\mathbf{k}_j$. The selection probabilities of each prompt are obtained via a softmax and their entropy is computed as:
\begin{equation}
p_j = \frac{\exp(s_j)}{\sum_{m=1}^{P}\exp(s_m)}, \quad H(\mathbf{p}) = - \sum_{j=1}^P p_j \log p_j,
\end{equation}
where $H(\mathbf{p})$ denotes the entropy of the probability distribution. Then the balance loss is defined as:
\begin{equation}
\mathcal{L}_{\mathrm{bal}} = \log P - H(\mathbf{p}).
\end{equation}
$\mathcal{L}_{\mathrm{bal}}$ encourages balanced prompt utilization during training. In addition, to enhance the sensitivity of instance-level prompt selection, we apply a contrastive regularization $\mathcal{L}_{con}$ as detailed in the Appendix~\ref{details}. All regularization terms are applied to both the two prompt pools.

\subsection{Overall Optimization Objective}
\label{Overall Optimization Objective}

The final training objective combines the primary reconstruction loss with cross-modal alignment loss and prompt regularization terms, and the total loss can be formulated as follows:

\begin{equation}
\mathcal{L} =
\underbrace{
    \lambda_{pix} \mathcal{L}_{pix}
    + \lambda_{fft} \mathcal{L}_{fft}
}_{\text{Reconstruction Loss}}
+
\underbrace{
    \lambda_{align} \mathcal{L}_{align}
    + \lambda_{div} \mathcal{L}_{div}
    + \lambda_{bal} \mathcal{L}_{bal}
    + \lambda_{con} \mathcal{L}_{con},
    % + \lambda_{orth} \mathcal{L}_{orth}
}_{\text{Cross-Modal Alignment and Prompt Regularization}}
\end{equation}
where $\mathcal{L}_{pix}$ and $\mathcal{L}_{fft}$ are $\ell_1$ loss in the RGB and Fourier domain respectively and $\lambda_{pix}$, $\lambda_{fft}$, $\lambda_{align}$, $\lambda_{div}$, $\lambda_{bal}$, and $\lambda_{con}$ are hyperparameters controlling the relative importance of each loss component.

\section{EXPERIMENTS}
\label{EXPERIMENTS}

\subsection{Experimental Setup}
To demonstrate the effectiveness of our method, we conduct experiments mainly from the 2 aspects: (1) 6-task and 3-domain experiment, (2) 9-task and 3-domain experiment. We consider 3 image domains—natural, medical, and remote sensing images—with a diverse selection of image restoration tasks from each domain. For the 6-task setting, we include 2 tasks per domain: natural image 4× super-resolution (SR) and deraining, medical MRI SR and CT denoising, and remote sensing image 4× SR and cloud removal. For the 9-task setting, we introduce one additional task per domain: natural image motion deblurring, medical PET synthesis, and remote sensing image dehazing. 

\textbf{Datasets and Evaluation Metrics.} The training datasets for each task are as follows: Natural image SR is trained on DF2K~\citep{DIV2K,F2K} dataset (DIV2K + Flickr2K) with 4× bicubic downsampling. Natural image deraining is trained using Rain100L~\citep{Rain100L}. Natural image deblurring uses the GoPro~\citep{GoPro} dataset. Following~\citet{amir,Restore-rwkv-allinone-medical}, medical MRI SR is trained on the IXI MRI dataset. Medical CT denoising uses dataset from the 2016 NIH AAPM-Mayo Clinic Low-Dose CT Grand Challenge~\citep{AAPM-CT-DN}. Medical PET synthesis is trained on the PolarStar m660 dataset, where both low-quality (LQ) and high-quality (HQ) PET images are reconstructed via the standard OSEM~\citep{osem-pet} method. Remote sensing image SR is trained on the UCMerced Land Use~\citep{UCMerced} dataset with 4× bicubic downsampling. Remote sensing cloud removal is trained using CUHK CR1~\citep{cuhk_cr1} dataset. Remote sensing dehazing is trained on RICE1~\citep{RICE}, which provides hazy and clean image pairs. Data augmentation including random cropping, horizontal flipping, and rotation are applied to improve robustness. Evaluation is performed on the corresponding test sets, using PSNR and SSIM~\citep{SSIM} in RGB space as the primary metrics.

\begin{table*}
    \centering
    \Large
    \caption{Quantitative comparison between our method and other SOTA methods on 3 domains \& 6 tasks experimental setting. The best and second-best metrics are highlighted in \textbf{bold} and \underline{underline}.}  
    \resizebox{\textwidth}{!}{
    \centering
    \begin{tabular}{cc|cccccccccccc|cc}
    \toprule
     \multicolumn{2}{c|}{\multirow{1}{*}{Image Domain}} & \multicolumn{4}{c}{Natural Image} & \multicolumn{4}{c}{Medical Image} & \multicolumn{4}{c}{Remote Sensing Image}&\multicolumn{2}{|c}{\multirow{3}{*}{\makecell{Average \\ Performance}}}  \\
    \cmidrule(lr){1-14}
     \multicolumn{2}{c|}{\multirow{2}{*}{Task \& Dataset}} & \multicolumn{2}{c}{Super-Resolution} & \multicolumn{2}{c}{Deraining} & \multicolumn{2}{c}{MRI SR} & \multicolumn{2}{c}{CT Denoising}& \multicolumn{2}{c}{RSI SR} & \multicolumn{2}{c|}{Cloud Removal}  \\
      && \multicolumn{2}{c}{on DIV2K-VAL} & \multicolumn{2}{c}{on Rain100L} & \multicolumn{2}{c}{on IXI MRI} & \multicolumn{2}{c}{on AAPM-Mayo}& \multicolumn{2}{c}{on UCMerced} & \multicolumn{2}{c|}{on CUHK CR1}  \\
    \cmidrule(lr){1-16}
    % \cmidrule(lr){3-8}
      \multicolumn{1}{c}{\multirow{1}{*}{Method}} &\multicolumn{1}{c}{\multirow{1}{*}{Year}} & PSNR $\uparrow$ & SSIM $\uparrow$ & PSNR $\uparrow$ & SSIM $\uparrow$& PSNR $\uparrow$ & SSIM $\uparrow$& PSNR $\uparrow$ & SSIM $\uparrow$& PSNR $\uparrow$ & SSIM $\uparrow$& PSNR $\uparrow$ & SSIM $\uparrow$& PSNR $\uparrow$ & SSIM $\uparrow$   \\
    \midrule
    \multicolumn{2}{c}{Single-Task Method}  & \\
    \midrule
    MPRNet& CVPR2021 & 28.82 &  0.8115 & 38.07 &0.9817& 26.84& 0.8891&33.60 & 0.9259 & 27.70 & 0.7730 & 25.35 & 0.7389& 30.06	&0.8534\\
    SwinIR& ICCVW2021 &  28.61 &  0.8051 & 36.07& 0.9736& 26.06&0.8766 &33.51 &0.9243  & 27.29 &0.7545 &24.36 &0.6552& 29.32	&0.8482\\
    % \rowcolor{gray!10}
    Restormer& CVPR2022 & \underline{28.94} & \underline{0.8158} & 38.34 & 0.9822 &27.58 & 0.9017&33.69 & 0.9268 & 28.01 & \underline{0.7844} &25.96 &0.7541 & \underline{30.42}	& 0.8608 \\
      NAFNet & ECCV2022  &28.73 & 0.8146 &37.06& 0.9773& 27.32& 0.8980&33.68 &0.9270 & 27.87 &0.7808 &25.99 &0.7591& 30.11 & 0.8595\\
      \midrule
    \multicolumn{2}{c}{All-in-One Method}  & \\
    \midrule
     Transweather & CVPR2022 & 27.40 & 0.7643 &33.20&0.9495& 24.59&0.8181 &31.98&0.9040& 25.97 &  0.6933& 22.95 & 0.5732& 27.68&0.7837\\
      PromptIR & NeurIPS2023  &28.77 & 0.8160 & \underline{38.71} & 0.9831 &27.61 &0.9023&33.71 & 0.9270 & 28.05 &0.7860& 25.81 &0.7518& 30.44 & 0.8610\\
      AMIR & MICCAI2024 &28.78 & 0.8139 & 38.10 & 0.9820 & 26.30& 0.8793& 33.66& 0.9262& 27.87 & 0.7797 & 25.75 & 0.7474& 30.08 & 0.8548 \\
             DFPIR & CVPR2025 & 27.69 & 0.7845 &37.33&0.9745& 24.59&0.8181 & 32.66 & 0.9137 & 25.97 &  0.6933& 26.02 & 0.7072& 29.04&	0.8152\\
     AdaIR & ICLR2025 & 28.81 & 0.8157 & 38.19 & 0.9816 & 27.54 & 0.9009 & 33.68& 0.9266 & 27.99 & 0.7840& 26.03 &0.7578& 30.37	&0.8611 \\
      MoCEIR & CVPR2025 & 28.16 & 0.8156 & 38.64 & \underline{0.9840}& \underline{27.75}& \underline{0.9027}& \underline{33.74}&\textbf{ 0.9278 }& \underline{28.06}& 0.7843 &\underline{26.06} & \textbf{0.7615} & 30.40&	\underline{0.8627}\\
      \midrule
        \multicolumn{5}{l}{Muti-Domain All-in-One Method}  & \\
    \midrule
    \rowcolor{yellow!8}
       \multicolumn{2}{c}{DATPRL-IR-\textbf{\textcolor{red}{6T}} \ \  (Ours)} & \textbf{28.98} & \textbf{0.8191} &  \textbf{39.56} & \textbf{0.9865} & \textbf{27.88} & \textbf{0.9053} & \textbf{33.80} & \textbf{0.9278} & \textbf{28.29} & \textbf{0.7917} & \textbf{26.12} & \underline{0.7612} &\textbf{30.77}	&\textbf{0.8653} \\
       \midrule
       \rowcolor{blue!5}
      \multicolumn{2}{c}{DATPRL-IR-\textbf{\textcolor{red}{7T}} \ \  (Ours)}& 29.03 & 0.8183 & 39.65 & 0.9866 & 27.78 & 0.9037 & 33.76 & 0.9269 & 28.28 & 0.7908 & 25.91 & 0.7594 & 30.74& 0.8643	\\
      \rowcolor{blue!5}
      \multicolumn{2}{c}{DATPRL-IR-\textbf{\textcolor{red}{8T}} \ \  (Ours)}& 28.99 & 0.8188 & 39.64 & 0.9866 & 27.82 & 0.9047 & 33.77 & 0.9269 & 28.31 & 0.7920 & 25.92 & 0.7590 & 30.74& 0.8647\\
      \rowcolor{blue!5}
      \multicolumn{2}{c}{DATPRL-IR-\textbf{\textcolor{red}{9T}} \ \  (Ours)}& 29.05 & 0.8181 & 39.67 & 0.9867  & 27.86 & 0.9045 & 33.77 & 0.9273  & 28.31 & 0.7913 & 26.00 & 0.7592 & 30.78& 0.8645\\
    \bottomrule
    \end{tabular}
    }
\label{table: 3 domains 6 tasks}
\end{table*}

\begin{figure}[t]
  \centering
  \setlength{\abovecaptionskip}{0.08cm}
  \includegraphics[width=1.0\textwidth]{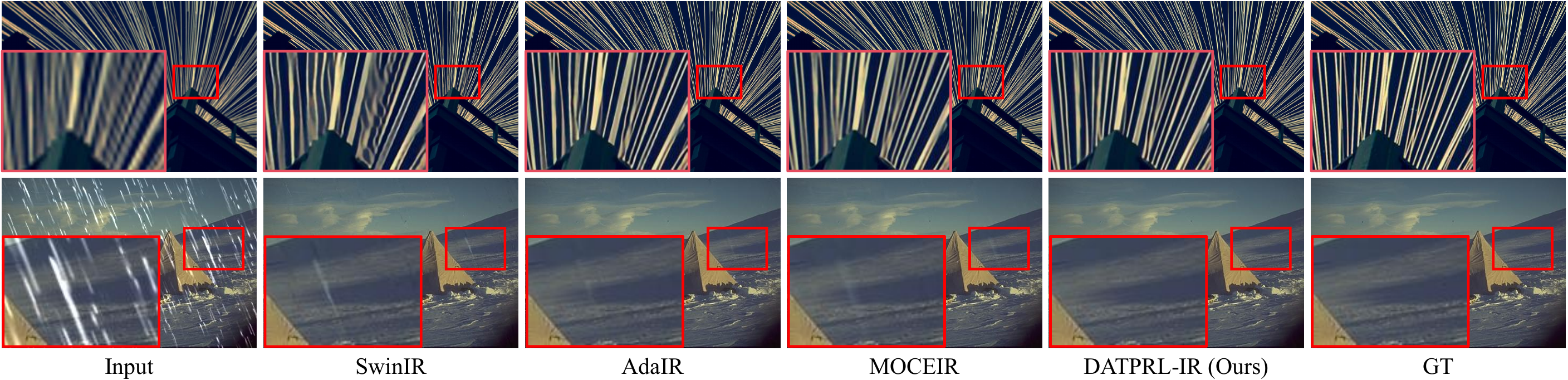}
  \caption{Comparison of our DATPRL-IR with other SOTA methods on 6-task and 3-domain setting.}
  \label{fig.Visual_Compare1}
\end{figure}

\textbf{Implementation Details.} We train our model under PyTorch~\citep{pytorch} framework using the Adam~\citep{adam} optimizer with $\beta_{1} = 0.9$, $\beta_{2} = 0.99$. The learning rate is initialized at $4 \times 10^{-4}$ with cosine annealing. Batch size is set to 12, and we train for 1000K iterations on NVIDIA RTX 5090 GPUs. We set the diversity threshold $\tau_{\text{div}} = 0.1$. The loss weights are set to $\lambda_{\text{align}} = 1.0$, $\lambda_{\text{div}} = 0.1$, $\lambda_{\text{con}} = 0.1$, and $\lambda_{\text{bal}} = 0.1$. For each prompt, the key is defined as a 1×1024 vector, while the value is set to 2×1024. The numbers of prompts in both the task and domain prompt pools are set to 15, with top-k selection configured as k=3 for the task prompt pool and k=5 for the domain prompt pool. The projector is a 3-layer lightweight CNN (mainly including Conv2d, AdaptiveAvgPool2d, and MLP), and the dimensionality of its output is 1024. To ensure a fair comparison, all competing methods are trained using the loss functions and specific training strategies adopted in their original papers, while all other training setting are kept the same as those used in training our model. \textbf{For more description on datasets and implementation details, please refer to Appendix \ref{details}.}

\subsection{multi-domain all-in-one image restoration}
\textbf{Results on 6-task and 3-domain all-in-one image restoration}\textbf{.}
To validate the effectiveness of our approach, we compare it with several SOTA AiOIR methods~\citep{moceir, adair, tian, amir, promptir, transweather} and classic image restoration baselines~\citep{NAFNet, restormer, SWINIR, MPRNet}. As shown in Table \ref{table: 3 domains 6 tasks}, our DATPRL-IR achieves almost comprehensive superiority across all six tasks, with an average PSNR improvement of 0.37 dB over the SOTA MoCEIR, and nearly 1 dB gain on the natural image deraining task. Furthermore, as illustrated in Figure \ref{fig.Visual_Compare1}, our method is able to more thoroughly remove degradations and reconstruct clearer image details compared to other methods. These results convincingly demonstrate the effectiveness of our proposed domain-aware task prompt representation learning in guiding image restoration.

\begin{figure}[t]
  \centering
  \setlength{\abovecaptionskip}{0.08cm}
  \includegraphics[width=1.0\textwidth]{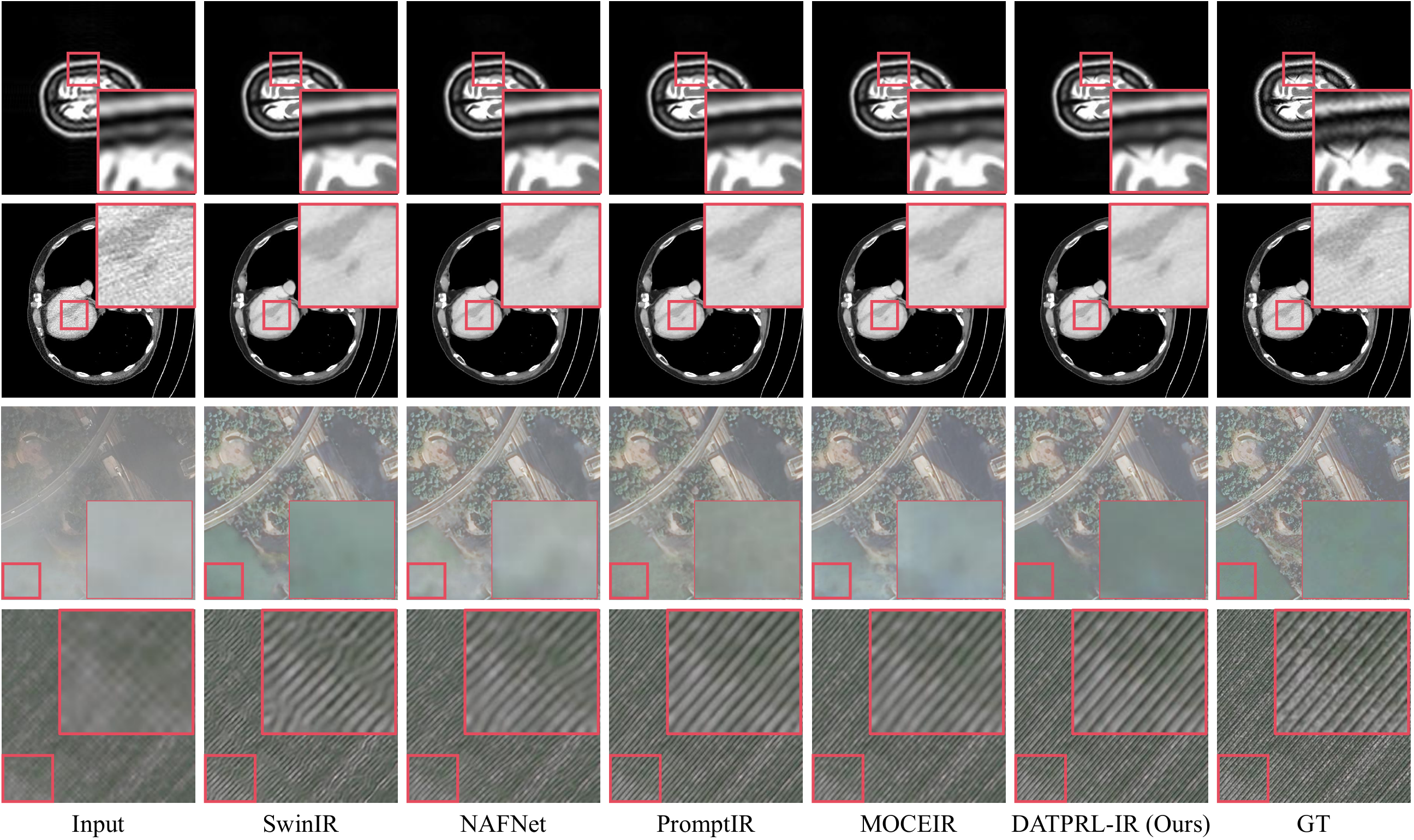}
  \caption{Comparison of our DATPRL-IR with other SOTA methods on 9-task and 3-domain setting.}
  \label{fig.Visual_Compare2}
\end{figure}

\textbf{Results on 9-task and 3-domain all-in-one image restoration.} To further evaluate scalability of our DATPRL-IR, we sequentially add three tasks—natural image deblurring, PET synthesis, and remote sensing image dehazing—to train our 9-task (9T) model, while also obtaining intermediate 7-task (7T) and 8-task (8T) models. As shown in Table \ref{table: 3 domains 6 tasks}, it can be clearly observed that when the task number grows from 6 to 9, our method does not exhibit significant performance degradation on the original tasks, it even achieves a certain degree of performance improvement. This provides strong evidence for our claim that different tasks indeed share transferable knowledge that can complement each other, and our method can effectively exploit both the shared and specific knowledge to enhance model robustness when facing a larger number of tasks. As illustrated in Figure \ref{fig.Visual_Compare2}, images restored by our method exhibit clearer textures and fewer artifacts. 

\textbf{Due to the limited space, additional quantitative and qualitative results, zero-shot and generalization results and analysis are provided in Appendix~\ref{additional_results}.} 

\subsection{Ablation Studies}
\textbf{The effects of task prompt pool and domain prompt pool.} Table \ref{table.The effects of task prompt pool and domain prompt pool} investigates the impact of task prompt (TP) and domain prompt (DP) pools under 6-task and 3-domain experiment setting. Using any one type of prompt pool can provide clear improvements over the baseline, indicating that both task-aware and domain-aware representations contribute useful prior knowledge. Importantly, combining both the two pools leads to the best performance, the PSNR on the three datasets is respectively increased by 1.22dB, 0.10dB, and 0.27dB across the three tasks. The above ablation results highlights the effectiveness of our dual-prompt design in enhancing generalization and robustness for scenarios with diverse tasks and domains.

\begin{minipage}[t]{0.6\textwidth}
\centering
\Large
% \begin{table}[h]
\scriptsize
  \centering
          \captionof{table}{Effect of task prompt (TP) pool and domain prompt (DP) pool on DATPRL-IR.}
\resizebox{\textwidth}{!}{
    \begin{tabular}{cc|cc|cc|cc}
    % \Large
    % \centering
    \toprule
     \multirow{3}{*}{TP Pool}  & \multirow{3}{*}{DP Pool} & \multicolumn{2}{c|}{Deraining} & \multicolumn{2}{c|}{CT Denoising}& \multicolumn{2}{c}{RSI SR} \\
      & & \multicolumn{2}{c|}{on Rain100L}  & \multicolumn{2}{c|}{on AAPM}& \multicolumn{2}{c}{on UCMerced}   \\
   && PSNR & SSIM & PSNR & SSIM & PSNR & SSIM\\
    \midrule
    \XSolidBrush & \XSolidBrush & 38.34 & 0.9823  & 33.70 & 0.9269 & 28.02 & 0.7844 \\
    \Checkmark & \XSolidBrush & 39.32 & 0.9855 & 33.76 & 0.9282 & 28.16 &0.7901\\
    \XSolidBrush &  \Checkmark & 38.88 & 0.9850 & 33.74 & 0.9268 & 28.12 &0.7897\\
   \rowcolor{blue!8}
  \Checkmark & \Checkmark &\textbf{39.56} & \textbf{0.9865} & \textbf{33.80} & \textbf{0.9278 } & \textbf{28.29} &\textbf{0.7917} \\
    \bottomrule
    \end{tabular} 
    }
  \label{table.The effects of task prompt pool and domain prompt pool}
% \end{table}
\end{minipage}
\begin{minipage}[t]{0.37\textwidth}
\Large
\centering
% \begin{table}[h]
\scriptsize
  \centering
          \captionof{table}{Effect of prompt numbers and top-k selection.}
\resizebox{\textwidth}{!}{
    \begin{tabular}{cc|cc|cc}
    % \centering
    \toprule
     \multicolumn{2}{c|}{\multirow{2}{*}{TP Pool}}  & \multicolumn{2}{c|}{\multirow{2}{*}{DP Pool}} & \multicolumn{2}{c}{6-Task Aver.}  \\
      & & & & \multicolumn{2}{c}{Performance}   \\
  Nums & Top K & Nums & Top K & PSNR & SSIM  \\
    \midrule
    10 & 1 & 10 & 1 &30.44 & 0.8607   \\
    10 & 3 & 10 & 5 &30.53 & 0.8612   \\
    15 & 1 & 15 & 1 &30.48 & 0.8607   \\
    % 10 & 3 & 10 & 3 &\\
     \rowcolor{blue!8}
    15 & 3 & 15 & 5 &\textbf{30.77} & \textbf{0.8653}  \\
    20 & 3 & 20 & 5 &30.73 & 0.8635  \\
  20 & 5 & 20 & 5 & 30.70 & 0.8638  \\
    \bottomrule
    \end{tabular} 
    }
  \label{table.The effects of prompt numbers and top-k selection}
% \end{table}
\end{minipage}

\textbf{The effects of prompt numbers and top-k selection.} As shown in Table \ref{table.The effects of prompt numbers and top-k selection}, we present the average performance over six tasks under different configurations of prompt numbers and top-k selection. The results reveal two trends. First, enlarging the prompt pool to a moderate size improves performance by offering richer choices, while excessively large pools lead to diminishing or even negative returns, as redundant prompts may dilute useful signals. Second, for top-k selection, using too few prompts limits expressiveness, while selecting too many reduces specificity. A balanced configuration not only preserves specificity but also better leverages the shared knowledge across tasks and domains. Overall, these results indicate that a moderate prompt capacity with carefully chosen retrieval breadth are key to achieving robust and generalized image restoration performance.

\textbf{The effects of different MLLMs.} To explore our method's sensitivity to the choice of MLLMs. We test our model by replacing LLaVA1.5-7B~\citep{LLaVa1.5} with two different MLLMs, LLaVA1.5-13B~\citep{LLaVa1.5} and Qwen3-VL-2B-Instruct~\citep{Qwen3, Qwen2.5-vl}. As shown in Table \ref{table: Effectiveness on different MLLMs}, the performance of our method remained stable under MLLM models of different parameter scales, with only marginal differences, demonstrating our method is not strongly dependent on a specific MLLM. This is because we only rely on relatively coarse domain-level semantics (e.g., “main image content”, “shooting view”, “brightness”, “color” etc.), and any MLLM capable of providing such descriptions is sufficient. 

\textbf{The effects of different prompt designs.}
To explore the effectiveness on different prompt designs, we replace our domain prompt pool with one fixed explicit text prompt per domain (e.g., “This is an MRI image”, “This is a natural image”) and simpler domain encodings (one learnable 2$\times$1024 tensor per domain). As shown in Table \ref{table: Effectiveness on different prompt designs}, both alternatives underperform our domain prompt pool on almost all tasks. Our method does not rely on a single fixed prompt per domain, but instead implements “shared + specific” prior modeling and learns a set of diverse domain-aware priors, allowing instance-level adaptive selection rather than forcing all MRI (or CT / RSI / Natural) images to share one identical prompt. Furthermore, using explicit prompts would require the user to know the domain beforehand, which contradicts our blind-restoration setting. 

\begin{table*}[h]
    \centering
    \Large
    % \captionsetup{skip=5pt}
    \caption{Effectiveness on different MLLMs on 3 domains \& 9 tasks experimental setting. The best metrics are highlighted in \textbf{bold}.}  
    \resizebox{\textwidth}{!}{
    \centering
    \begin{tabular}{c|cccccccccccc}
    \toprule
    \cmidrule(lr){1-13}
     \multicolumn{1}{c|}{\multirow{2}{*}{Task \& Dataset}} & \multicolumn{2}{c}{Natural SR} & \multicolumn{2}{c}{Deraining}  & \multicolumn{2}{c}{CT Denoising} & \multicolumn{2}{c}{PET Synthesis} & \multicolumn{2}{c}{RSI SR} & \multicolumn{2}{c}{RSI Dehazing} \\
   & \multicolumn{2}{c}{on DIV2K-Val} & \multicolumn{2}{c}{on Rain100L}  & \multicolumn{2}{c}{on AAPM} & \multicolumn{2}{c}{on PolarStar} & \multicolumn{2}{c}{on UCMerced} & \multicolumn{2}{c}{on RICE1}  \\
    \cmidrule(lr){1-13}
    % \cmidrule(lr){3-8}
      \multicolumn{1}{c}{\multirow{1}{*}{Method}}  & PSNR $\uparrow$ & SSIM $\uparrow$ & PSNR $\uparrow$ & SSIM $\uparrow$& PSNR $\uparrow$& SSIM $\uparrow$& PSNR $\uparrow$ & SSIM $\uparrow$ & PSNR $\uparrow$& SSIM $\uparrow$& PSNR $\uparrow$ & SSIM $\uparrow$\\

    \midrule
      LLaVA1.5-13B  &  29.04 & 0.8180  &\textbf{ 39.71} & \textbf{0.9868 }& \textbf{33.77} & 0.9272 & 37.08 & 0.9500 & \textbf{28.31} & \textbf{0.7918} & 26.92 & 0.9347\\
      \rowcolor{blue!8}
       LLaVA1.5-7B & 29.05 & 0.8181 & 39.67 & 0.9867  & \textbf{33.77} & \textbf{0.9273} &  \textbf{37.12} & \textbf{0.9502} & \textbf{28.31} & 0.7913& \textbf{26.94} & 0.9347 \\
      Qwen3-VL-2B-Instruct   & \textbf{29.07} & \textbf{0.8182} & 39.69 & \textbf{0.9868} & \textbf{33.77} & 0.9270 & 37.11 & 0.9501 & \textbf{28.31} & 0.7914 & 26.82 &   \textbf{0.9351}\\
    \bottomrule
    \end{tabular}
    }
    % \end{center}
\label{table: Effectiveness on different MLLMs}
\end{table*}

\begin{table*}[h]
    \centering
    \Large
    \caption{Effectiveness on different prompt designs on 3 domains \& 9 tasks experimental setting. The best metrics are highlighted in \textbf{bold}.}
    \resizebox{\textwidth}{!}{
    \centering
    \begin{tabular}{c|cccccccccccc}
    \toprule
    \cmidrule(lr){1-13}
     \multicolumn{1}{c|}{\multirow{2}{*}{Task \& Dataset}} & \multicolumn{2}{c}{Deraining}  & \multicolumn{2}{c}{Deblurring} & \multicolumn{2}{c}{MRI SR} & \multicolumn{2}{c}{PET Synthesis} & \multicolumn{2}{c}{Cloud Removal} & \multicolumn{2}{c}{RSI Dehazing} \\
   & \multicolumn{2}{c}{on Rain100L}  & \multicolumn{2}{c}{on GoPro}  & \multicolumn{2}{c}{on IXI MRI} & \multicolumn{2}{c}{on PolarStar} & \multicolumn{2}{c}{on CUHK CR1} & \multicolumn{2}{c}{on RICE1}  \\
    \cmidrule(lr){1-13}
      \multicolumn{1}{c}{\multirow{1}{*}{Method}}  & PSNR $\uparrow$ & SSIM $\uparrow$ & PSNR $\uparrow$ & SSIM $\uparrow$& PSNR $\uparrow$& SSIM $\uparrow$& PSNR $\uparrow$ & SSIM $\uparrow$ & PSNR $\uparrow$& SSIM $\uparrow$& PSNR $\uparrow$ & SSIM $\uparrow$\\
    \midrule
     Explicit Domain Prompts  & 39.59 & 0.9865 & 29.31 & 0.8839 & 27.77 & 0.9035 & 37.08 & 0.9500 & 25.76 & 0.7573 &25.91 &0.9296\\
      Simple Domain Encodings &39.52 & 0.9863  &29.21 & 0.8814 & 27.70 & 0.9031 & 37.02 & 0.9496 & 25.81 & 0.7566 &25.81 &0.9213\\
      \rowcolor{blue!8}
      Domain Prompt Pool \  (Ours) & \textbf{39.67} & \textbf{0.9867} & \textbf{29.57} & \textbf{0.8881} & \textbf{27.86} & \textbf{0.9045} & \textbf{37.12} & \textbf{0.9502} &  \textbf{26.00} & \textbf{0.7592} & \textbf{26.94} & \textbf{0.9347} \\
    \bottomrule
    \end{tabular}
    }
\label{table: Effectiveness on different prompt designs}
\end{table*}

\subsection{In-depth Analysis and Discussion.} 
\begin{figure}[t]
  \centering
  \setlength{\abovecaptionskip}{0.05cm}
  \begin{subfigure}[t]{1.0\textwidth}
  \centering
  \includegraphics[width=1.0\textwidth]{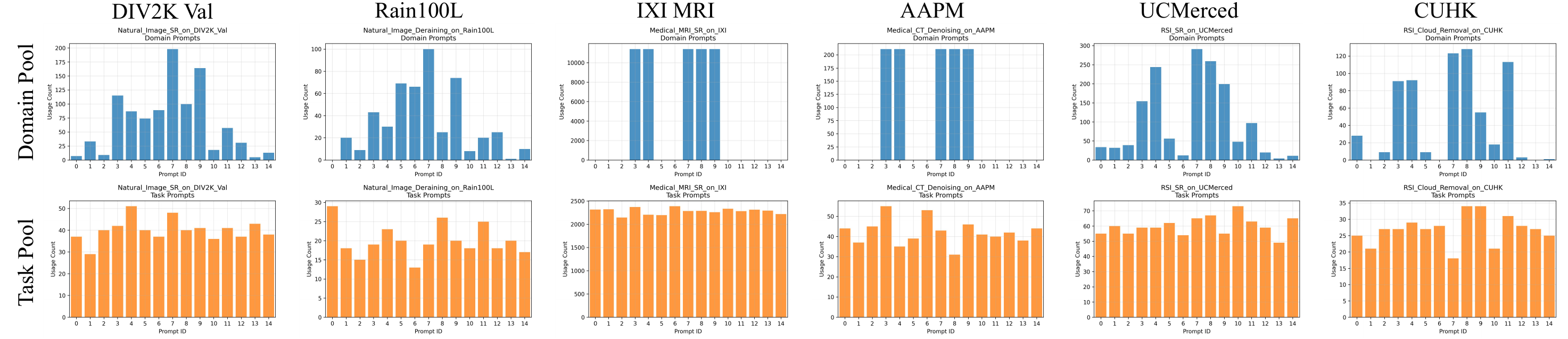}
  \caption{Prompt selection distribution of domain prompt pool and task prompt pool on partial test sets.}
  \label{fig.prompt_select}
\end{subfigure}
  \begin{subfigure}[t]{0.62\textwidth}
    \centering
    \includegraphics[width=\linewidth]{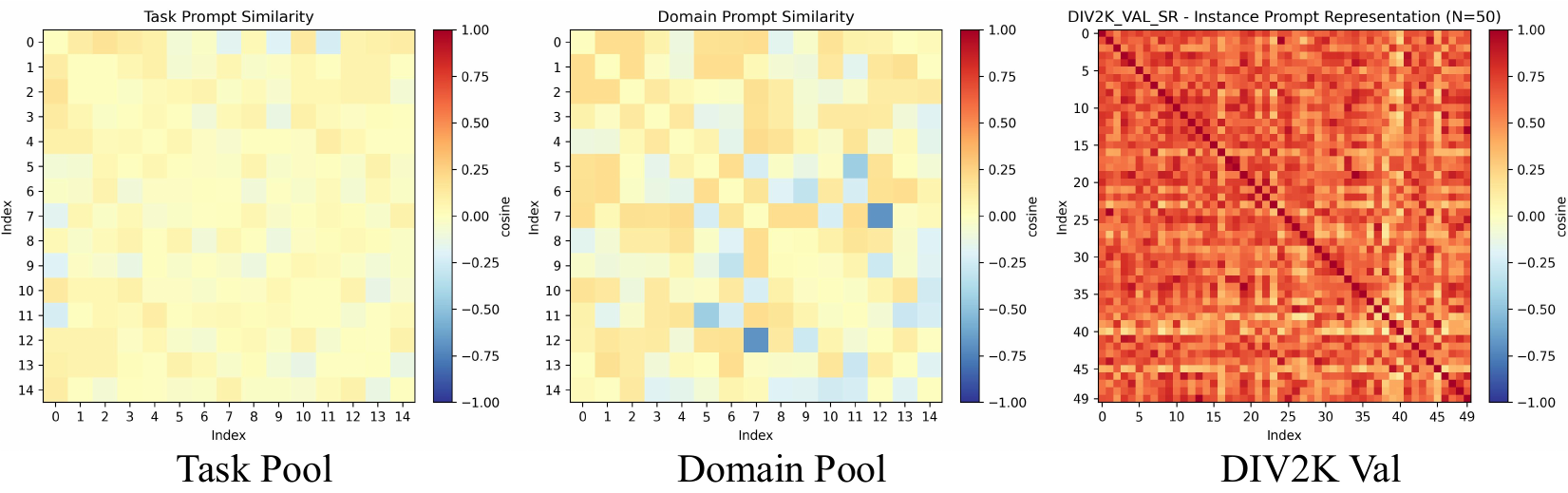}
    \caption{Diversity of the prompts in the 2 prompt pool and instance-level prompt representations on DIV2K-Val.}
    \label{fig.prompt_similarity}
  \end{subfigure}
  \hfill
  \begin{subfigure}[t]{0.37\textwidth}
    \centering
    \includegraphics[width=\linewidth]{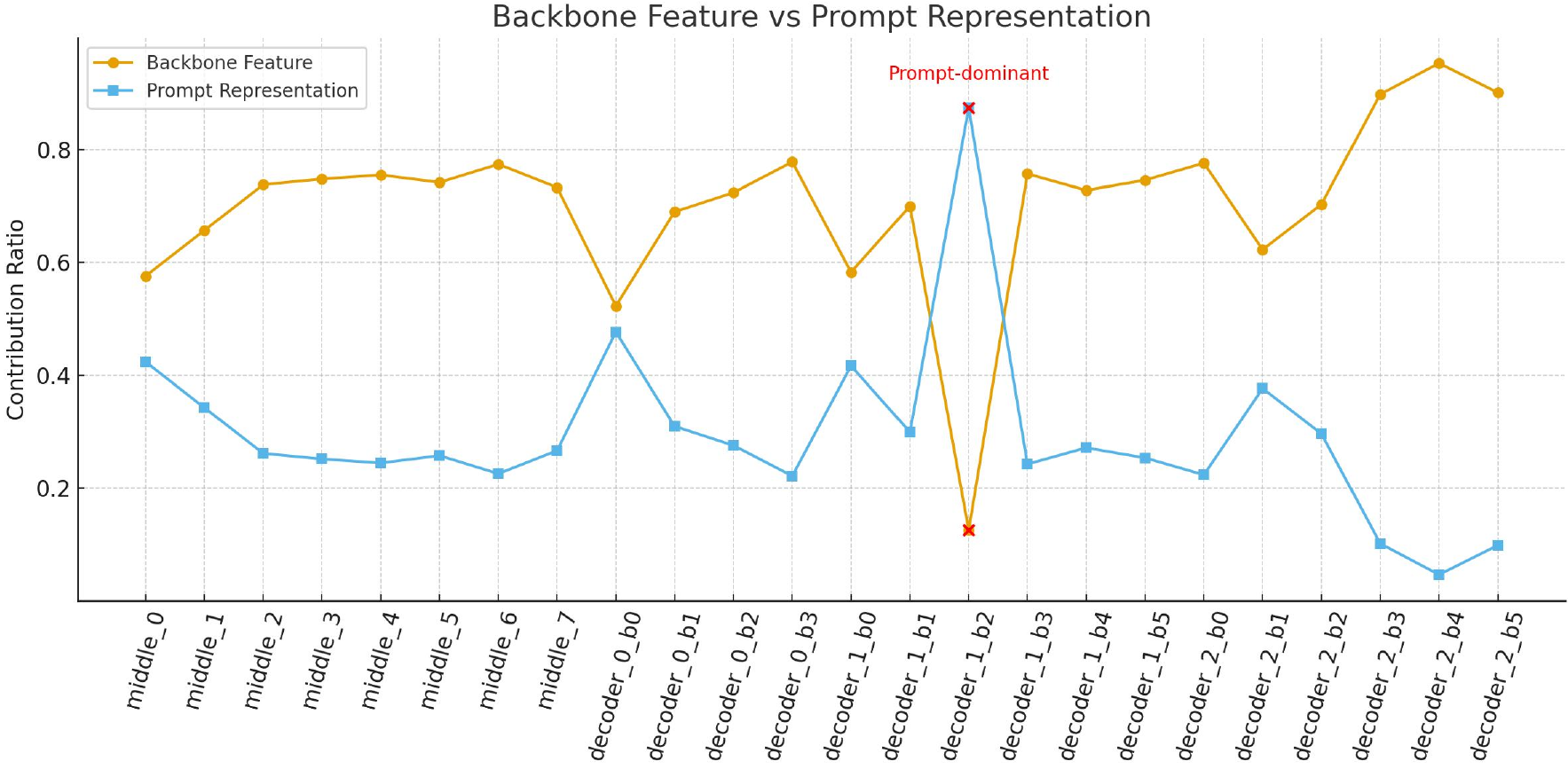}
    \caption{Contribution ratio of prompt representations and backbone features.}
    \label{fig.contribution_ratio}
  \end{subfigure}
  \caption{In-depth analysis of our method. Zoom in for better visualization.}
  \label{fig.prompt_analysis}
\end{figure}

\textbf{Prompt selection distribution.} As shown in Figure \ref{fig.prompt_select}, we visualize the selection distribution of prompts from both the domain and task prompt pools across six datasets. For the domain prompt pool, different datasets exhibit distinct distributions. Notably, medical datasets present highly uniform selections due to their relatively homogeneous image content and color patterns. In contrast, the task prompt pool shows larger overlaps across datasets, suggesting that a considerable number of task prompts are shared. This observation confirms that our method can effectively leverage shared knowledge across tasks to enhance network performance.

\textbf{Prompt diversity.} Figure \ref{fig.prompt_similarity} presents pairwise similarity heatmaps of prompt values in the two prompt pools. It is evident that each prompt has learned distinct content with little redundancy, thereby providing the model with diverse options. Furthermore, to examine instance-level behavior, we visualize the similarities between final prompt representations generated for different input images from the DIV2K validation set. For the same task, our method produces prompt representations with similar overall directions, while retaining instance-specific variations, indicating that our method enhances the instance-level diversity of learned prompt representations.

\textbf{The contribution ratio of prompt representations and backbone features.} We further analyze the learnable gating coefficient $\alpha_l \in [0,1]$ between prompt representations and backbone features at each block from the middle layers to the decoder, which is mentioned in Sec.~\ref{Domain-Aware Task Prompt Representation Learning}. As shown in Figure \ref{fig.contribution_ratio}, most blocks exhibit a dominant reliance on backbone features, indicating that the network is still primarily driven by the restoration backbone while prompt representations serve as auxiliary guidance. Furthermore, earlier blocks at each scale rely more heavily on prompt representations compared to deeper blocks, and  the large variations in contribution ratios across different layers further highlight the importance of adopting adaptive fusion ratios.

\section{Conclusion}
\label{conclusion}

In this work, we proposed the first multi-domain all-in-one image restoration (MD-AiOIR) method, DATPRL-IR, which covers multiple restoration tasks across various image domains. By introducing domain-aware task representation learning, DATPRL-IR can fully utilize both specific and shared knowledge across tasks and domains, effectively reducing the learning difficulty of the model and improving its performance. Extensive experiments show that DATPRL-IR outperforms existing SOTA methods and demonstrates excellent generalization abilities. We believe that this work lays the foundation for future research towards a more unified restoration framework.

\section*{Ethics Statement}
The authors acknowledge that this work adheres to the ICLR Code of Ethics.
% \section*{Reproducibility statement} The code, trained models and datasets to reproduce our experiments will be available after this paper is accepted.


\bibliography{iclr2026_conference}

@article{CNN,
  title={Deep learning},
  author={LeCun, Yann and Bengio, Yoshua and Hinton, Geoffrey},
  journal={nature},
  volume={521},
  number={7553},
  pages={436--444},
  year={2015},
  publisher={Nature Publishing Group UK London}
}

@article{Transformer,
  title={Attention is all you need},
  author={Vaswani, Ashish and Shazeer, Noam and Parmar, Niki and Uszkoreit, Jakob and Jones, Llion and Gomez, Aidan N and Kaiser, {\L}ukasz and Polosukhin, Illia},
  journal={Advances in neural information processing systems},
  volume={30},
  year={2017}
}

@article{ViT,
  title={An image is worth 16x16 words: Transformers for image recognition at scale},
  author={Dosovitskiy, Alexey and Beyer, Lucas and Kolesnikov, Alexander and Weissenborn, Dirk and Zhai, Xiaohua and Unterthiner, Thomas and Dehghani, Mostafa and Minderer, Matthias and Heigold, Georg and Gelly, Sylvain and others},
  journal={arXiv preprint arXiv:2010.11929},
  year={2020}
}

@inproceedings{resnet,
  title={Deep residual learning for image recognition},
  author={He, Kaiming and Zhang, Xiangyu and Ren, Shaoqing and Sun, Jian},
  booktitle={Proceedings of the IEEE conference on computer vision and pattern recognition},
  pages={770--778},
  year={2016}
}

@article{Mamba,
  title={Mamba: Linear-time sequence modeling with selective state spaces},
  author={Gu, Albert and Dao, Tri},
  journal={arXiv preprint arXiv:2312.00752},
  year={2023}
}

@article{Vision_mamba,
  title={Vision mamba: Efficient visual representation learning with bidirectional state space model},
  author={Zhu, Lianghui and Liao, Bencheng and Zhang, Qian and Wang, Xinlong and Liu, Wenyu and Wang, Xinggang},
  journal={arXiv preprint arXiv:2401.09417},
  year={2024}
}

@inproceedings{SFD,
  title={Exploring semantic feature discrimination for perceptual image super-resolution and opinion-unaware no-reference image quality assessment},
  author={Dong, Guanglu and Liao, Xiangyu and Li, Mingyang and Guo, Guihuan and Ren, Chao},
  booktitle={Proceedings of the Computer Vision and Pattern Recognition Conference},
  pages={28176--28187},
  year={2025}
}

@inproceedings{CSUD,
  title={Channel Consistency Prior and Self-Reconstruction Strategy Based Unsupervised Image Deraining},
  author={Dong, Guanglu and Zheng, Tianheng and Cao, Yuanzhouhan and Qing, Linbo and Ren, Chao},
  booktitle={Proceedings of the Computer Vision and Pattern Recognition Conference},
  pages={7469--7479},
  year={2025}
}

@inproceedings{SWINIR,
  title={Swinir: Image restoration using swin transformer},
  author={Liang, Jingyun and Cao, Jiezhang and Sun, Guolei and Zhang, Kai and Van Gool, Luc and Timofte, Radu},
  booktitle={Proceedings of the IEEE/CVF international conference on computer vision},
  pages={1833--1844},
  year={2021}
}

@inproceedings{restormer,
  title={Restormer: Efficient transformer for high-resolution image restoration},
  author={Zamir, Syed Waqas and Arora, Aditya and Khan, Salman and Hayat, Munawar and Khan, Fahad Shahbaz and Yang, Ming-Hsuan},
  booktitle={Proceedings of the IEEE/CVF conference on computer vision and pattern recognition},
  pages={5728--5739},
  year={2022}
}

@article{promptir,
  title={Promptir: Prompting for all-in-one image restoration},
  author={Potlapalli, Vaishnav and Zamir, Syed Waqas and Khan, Salman H and Shahbaz Khan, Fahad},
  journal={Advances in Neural Information Processing Systems},
  volume={36},
  year={2023}
}

@inproceedings{NAFNet,
  title={Simple baselines for image restoration},
  author={Chen, Liangyu and Chu, Xiaojie and Zhang, Xiangyu and Sun, Jian},
  booktitle={European conference on computer vision},
  pages={17--33},
  year={2022},
  organization={Springer}
}

@article{pytorch,
  title={Pytorch: An imperative style, high-performance deep learning library},
  author={Paszke, Adam and Gross, Sam and Massa, Francisco and Lerer, Adam and Bradbury, James and Chanan, Gregory and Killeen, Trevor and Lin, Zeming and Gimelshein, Natalia and Antiga, Luca and others},
  journal={Advances in neural information processing systems},
  volume={32},
  year={2019}
}

@article{adam,
  title={Adam: A method for stochastic optimization},
  author={Kingma, Diederik P},
  journal={arXiv preprint arXiv:1412.6980},
  year={2014}
}

@inproceedings{MPRNet,
  title={Multi-stage progressive image restoration},
  author={Zamir, Syed Waqas and Arora, Aditya and Khan, Salman and Hayat, Munawar and Khan, Fahad Shahbaz and Yang, Ming-Hsuan and Shao, Ling},
  booktitle={Proceedings of the IEEE/CVF conference on computer vision and pattern recognition},
  pages={14821--14831},
  year={2021}
}

@inproceedings{mambair,
  title={Mambair: A simple baseline for image restoration with state-space model},
  author={Guo, Hang and Li, Jinmin and Dai, Tao and Ouyang, Zhihao and Ren, Xudong and Xia, Shu-Tao},
  booktitle={European conference on computer vision},
  pages={222--241},
  year={2024},
  organization={Springer}
}

@inproceedings{moceir,
  title={Complexity experts are task-discriminative learners for any image restoration},
  author={Zamfir, Eduard and Wu, Zongwei and Mehta, Nancy and Tan, Yuedong and Paudel, Danda Pani and Zhang, Yulun and Timofte, Radu},
  booktitle={Proceedings of the Computer Vision and Pattern Recognition Conference},
  pages={12753--12763},
  year={2025}
}

@inproceedings{transweather,
  title={Transweather: Transformer-based restoration of images degraded by adverse weather conditions},
  author={Valanarasu, Jeya Maria Jose and Yasarla, Rajeev and Patel, Vishal M},
  booktitle={Proceedings of the IEEE/CVF conference on computer vision and pattern recognition},
  pages={2353--2363},
  year={2022}
}

@article{adair,
  title={Adair: Adaptive all-in-one image restoration via frequency mining and modulation},
  author={Cui, Yuning and Zamir, Syed Waqas and Khan, Salman and Knoll, Alois and Shah, Mubarak and Khan, Fahad Shahbaz},
  journal={arXiv preprint arXiv:2403.14614},
  year={2024}
}

@inproceedings{amir,
  title={All-in-one medical image restoration via task-adaptive routing},
  author={Yang, Zhiwen and Chen, Haowei and Qian, Ziniu and Yi, Yang and Zhang, Hui and Zhao, Dan and Wei, Bingzheng and Xu, Yan},
  booktitle={International Conference on Medical Image Computing and Computer-Assisted Intervention},
  pages={67--77},
  year={2024},
  organization={Springer}
}

@article{Restore-rwkv-allinone-medical,
  title={Restore-rwkv: Efficient and effective medical image restoration with rwkv},
  author={Yang, Zhiwen and Li, Jiayin and Zhang, Hui and Zhao, Dan and Wei, Bingzheng and Xu, Yan},
  journal={arXiv preprint arXiv:2407.11087},
  year={2024}
}

@article{cuhk_cr1,
  title={Diffusion enhancement for cloud removal in ultra-resolution remote sensing imagery},
  author={Sui, Jialu and Ma, Yiyang and Yang, Wenhan and Zhang, Xiaokang and Pun, Man-On and Liu, Jiaying},
  journal={IEEE Transactions on Geoscience and Remote Sensing},
  year={2024},
  publisher={IEEE}
}

@inproceedings{UCMerced,
  title={Bag-of-visual-words and spatial extensions for land-use classification},
  author={Yang, Yi and Newsam, Shawn},
  booktitle={Proceedings of the 18th SIGSPATIAL international conference on advances in geographic information systems},
  pages={270--279},
  year={2010}
}

@article{AAPM-CT-DN,
  title={Low-dose CT for the detection and classification of metastatic liver lesions: results of the 2016 low dose CT grand challenge},
  author={McCollough, Cynthia H and Bartley, Adam C and Carter, Rickey E and Chen, Baiyu and Drees, Tammy A and Edwards, Phillip and Holmes III, David R and Huang, Alice E and Khan, Farhana and Leng, Shuai and others},
  journal={Medical physics},
  volume={44},
  number={10},
  pages={e339--e352},
  year={2017},
  publisher={Wiley Online Library}
}

@article{osem-pet,
  title={Accelerated image reconstruction using ordered subsets of projection data},
  author={Hudson, H Malcolm and Larkin, Richard S},
  journal={IEEE transactions on medical imaging},
  volume={13},
  number={4},
  pages={601--609},
  year={1994},
  publisher={IEEE}
}

@article{Rain100L,
  title={Joint rain detection and removal from a single image with contextualized deep networks},
  author={Yang, Wenhan and Tan, Robby T and Feng, Jiashi and Guo, Zongming and Yan, Shuicheng and Liu, Jiaying},
  journal={IEEE transactions on pattern analysis and machine intelligence},
  volume={42},
  number={6},
  pages={1377--1393},
  year={2019},
  publisher={IEEE}
}

@inproceedings{DIV2K,
  title={Ntire 2017 challenge on single image super-resolution: Dataset and study},
  author={Agustsson, Eirikur and Timofte, Radu},
  booktitle={Proceedings of the IEEE conference on computer vision and pattern recognition workshops},
  pages={126--135},
  year={2017}
}

@InProceedings{F2K,
author = {Timofte, Radu and Agustsson, Eirikur and Van Gool, Luc and Yang, Ming-Hsuan and Zhang, Lei},
title = {NTIRE 2017 Challenge on Single Image Super-Resolution: Methods and Results},
booktitle = {Proceedings of the IEEE Conference on Computer Vision and Pattern Recognition (CVPR) Workshops},
month = {July},
year = {2017}
}

@inproceedings{GoPro,
  title={Deep multi-scale convolutional neural network for dynamic scene deblurring},
  author={Nah, Seungjun and Hyun Kim, Tae and Mu Lee, Kyoung},
  booktitle={Proceedings of the IEEE conference on computer vision and pattern recognition},
  pages={3883--3891},
  year={2017}
}

@inproceedings{tian,
  title={Degradation-Aware Feature Perturbation for All-in-One Image Restoration},
  author={Tian, Xiangpeng and Liao, Xiangyu and Liu, Xiao and Li, Meng and Ren, Chao},
  booktitle={Proceedings of the Computer Vision and Pattern Recognition Conference},
  pages={28165--28175},
  year={2025}
}

@inproceedings{SPANet,
  title={Spatial attentive single-image deraining with a high quality real rain dataset},
  author={Wang, Tianyu and Yang, Xin and Xu, Ke and Chen, Shaozhe and Zhang, Qiang and Lau, Rynson WH},
  booktitle={Proceedings of the IEEE/CVF conference on computer vision and pattern recognition},
  pages={12270--12279},
  year={2019}
}

@ARTICLE{Perceive-IR,
  author={Zhang, Xu and Ma, Jiaqi and Wang, Guoli and Zhang, Qian and Zhang, Huan and Zhang, Lefei},
  journal={IEEE Transactions on Image Processing}, 
  title={Perceive-IR: Learning to Perceive Degradation Better for All-in-One Image Restoration}, 
  year={2025},
  doi={10.1109/TIP.2025.3566300}}

@InProceedings{LLaVa1.5,
    author    = {Liu, Haotian and Li, Chunyuan and Li, Yuheng and Lee, Yong Jae},
    title     = {Improved Baselines with Visual Instruction Tuning},
    booktitle = {Proceedings of the IEEE/CVF Conference on Computer Vision and Pattern Recognition (CVPR)},
    month     = {June},
    year      = {2024},
    pages     = {26296-26306}
}

@InProceedings{CLIP,
  title = 	 {Learning Transferable Visual Models From Natural Language Supervision},
  author =       {Radford, Alec and Kim, Jong Wook and Hallacy, Chris and Ramesh, Aditya and Goh, Gabriel and Agarwal, Sandhini and Sastry, Girish and Askell, Amanda and Mishkin, Pamela and Clark, Jack and Krueger, Gretchen and Sutskever, Ilya},
  booktitle = 	 {Proceedings of the 38th International Conference on Machine Learning},
  pages = 	 {8748--8763},
  year = 	 {2021},
  volume = 	 {139},
  month = 	 {18--24 Jul},
  publisher =    {PMLR},
}

@misc{vicuna,
    title = {Vicuna: An Open-Source Chatbot Impressing GPT-4 with 90\%* ChatGPT Quality},
    url = {https://lmsys.org/blog/2023-03-30-vicuna/},
    author = {Chiang, Wei-Lin and Li, Zhuohan and Lin, Zi and Sheng, Ying and Wu, Zhanghao and Zhang, Hao and Zheng, Lianmin and Zhuang, Siyuan and Zhuang, Yonghao and Gonzalez, Joseph E. and Stoica, Ion and Xing, Eric P.},
    month = {March},
    year = {2023}
}

@misc{LLaMA,
      title={LLaMA: Open and Efficient Foundation Language Models}, 
      author={Hugo Touvron and Thibaut Lavril and Gautier Izacard and Xavier Martinet and Marie-Anne Lachaux and Timothée Lacroix and Baptiste Rozière and Naman Goyal and Eric Hambro and Faisal Azhar and Aurelien Rodriguez and Armand Joulin and Edouard Grave and Guillaume Lample},
      year={2023},
      eprint={2302.13971},
      archivePrefix={arXiv},
      primaryClass={cs.CL},
      url={https://arxiv.org/abs/2302.13971}, 
}

@inproceedings{instructir,
  title={Instructir: High-quality image restoration following human instructions},
  author={Conde, Marcos V and Geigle, Gregor and Timofte, Radu},
  booktitle={European Conference on Computer Vision},
  pages={1--21},
  year={2024},
  organization={Springer}
}

@inproceedings{airnet,
  title={All-in-one image restoration for unknown corruption},
  author={Li, Boyun and Liu, Xiao and Hu, Peng and Wu, Zhongqin and Lv, Jiancheng and Peng, Xi},
  booktitle={Proceedings of the IEEE/CVF conference on computer vision and pattern recognition},
  pages={17452--17462},
  year={2022}
}

@article{all-in-one-medir2025,
  title={All-in-One Medical Image Restoration with Latent Diffusion-Enhanced Vector-Quantized Codebook Prior},
  author={Chen, Haowei and Yang, Zhiwen and Hou, Haotian and Zhang, Hui and Wei, Bingzheng and Zhou, Gang and Xu, Yan},
  journal={arXiv preprint arXiv:2507.19874},
  year={2025}
}

@article{all-in-one-medir-Restore-rwkv,
  title={Restore-rwkv: Efficient and effective medical image restoration with rwkv},
  author={Yang, Zhiwen and Li, Jiayin and Zhang, Hui and Zhao, Dan and Wei, Bingzheng and Xu, Yan},
  journal={IEEE Journal of Biomedical and Health Informatics},
  year={2025},
  publisher={IEEE}
}

@article{FrePrompter,
  title={FrePrompter: Frequency self-prompt for all-in-one image restoration},
  author={Wu, Zhijian and Liu, Wenhui and Wang, Jingchao and Li, Jun and Huang, Dingjiang},
  journal={Pattern Recognition},
  volume={161},
  pages={111223},
  year={2025},
  publisher={Elsevier}
}

@article{Prores,
  title={Prores: Exploring degradation-aware visual prompt for universal image restoration},
  author={Ma, Jiaqi and Cheng, Tianheng and Wang, Guoli and Zhang, Qian and Wang, Xinggang and Zhang, Lefei},
  journal={arXiv preprint arXiv:2306.13653},
  year={2023}
}

@article{all-in-one-Prompt,
  title={Towards effective multiple-in-one image restoration: A sequential and prompt learning strategy},
  author={Kong, Xiangtao and Dong, Chao and Zhang, Lei},
  journal={arXiv preprint arXiv:2401.03379},
  year={2024}
}

@article{Promptrestorer,
  title={Promptrestorer: A prompting image restoration method with degradation perception},
  author={Wang, Cong and Pan, Jinshan and Wang, Wei and Dong, Jiangxin and Wang, Mengzhu and Ju, Yakun and Chen, Junyang},
  journal={Advances in Neural Information Processing Systems},
  volume={36},
  pages={8898--8912},
  year={2023}
}

@inproceedings{L2P,
  title={Learning to prompt for continual learning},
  author={Wang, Zifeng and Zhang, Zizhao and Lee, Chen-Yu and Zhang, Han and Sun, Ruoxi and Ren, Xiaoqi and Su, Guolong and Perot, Vincent and Dy, Jennifer and Pfister, Tomas},
  booktitle={Proceedings of the IEEE/CVF conference on computer vision and pattern recognition},
  pages={139--149},
  year={2022}
}

@inproceedings{IDR,
  title={Ingredient-oriented multi-degradation learning for image restoration},
  author={Zhang, Jinghao and Huang, Jie and Yao, Mingde and Yang, Zizheng and Yu, Hu and Zhou, Man and Zhao, Feng},
  booktitle={Proceedings of the IEEE/CVF conference on computer vision and pattern recognition},
  pages={5825--5835},
  year={2023}
}

@article{DA-CLIP,
  title={Controlling vision-language models for universal image restoration},
  author={Luo, Ziwei and Gustafsson, Fredrik K and Zhao, Zheng and Sj{\"o}lund, Jens and Sch{\"o}n, Thomas B},
  journal={arXiv preprint arXiv:2310.01018},
  volume={3},
  number={8},
  year={2023}
}

@article{DCPT,
  title={Universal Image Restoration Pre-training via Degradation Classification},
  author={JiaKui Hu and Lujia Jin and Zhengjian Yao and Yanye Lu},
  journal={The Thirteenth International Conference on Learning Representations},
  year={2025}
}

@article{prompt-all-in-one-tcsvt,
  title={Prompt-based ingredient-oriented all-in-one image restoration},
  author={Gao, Hu and Yang, Jing and Zhang, Ying and Wang, Ning and Yang, Jingfan and Dang, Depeng},
  journal={IEEE Transactions on Circuits and Systems for Video Technology},
  volume={34},
  number={10},
  pages={9458--9471},
  year={2024},
  publisher={IEEE}
}

@inproceedings{MedIR_MRI_SR,
  title={Efficient and accurate MRI super-resolution using a generative adversarial network and 3D multi-level densely connected network},
  author={Chen, Yuhua and Shi, Feng and Christodoulou, Anthony G and Xie, Yibin and Zhou, Zhengwei and Li, Debiao},
  booktitle={International conference on medical image computing and computer-assisted intervention},
  pages={91--99},
  year={2018},
  organization={Springer}
}

@inproceedings{MedIR_CT_denoising,
  title={Low-dose CT denoising with convolutional neural network},
  author={Chen, Hu and Zhang, Yi and Zhang, Weihua and Liao, Peixi and Li, Ke and Zhou, Jiliu and Wang, Ge},
  booktitle={2017 IEEE 14th international symposium on biomedical imaging (ISBI 2017)},
  pages={143--146},
  year={2017},
  organization={IEEE}
}

@article{MedIR_PET_synthesis,
  title={Adaptive rectification based adversarial network with spectrum constraint for high-quality PET image synthesis},
  author={Luo, Yanmei and Zhou, Luping and Zhan, Bo and Fei, Yuchen and Zhou, Jiliu and Wang, Yan and Shen, Dinggang},
  journal={Medical image analysis},
  volume={77},
  pages={102335},
  year={2022},
  publisher={Elsevier}
}

@article{remote_sensing_IR1,
  title={Super-resolution for remote sensing images via local--global combined network},
  author={Lei, Sen and Shi, Zhenwei and Zou, Zhengxia},
  journal={IEEE Geoscience and Remote Sensing Letters},
  volume={14},
  number={8},
  pages={1243--1247},
  year={2017},
  publisher={IEEE}
}

@article{remote_sensing_IR2,
  title={Lightweight stepless super-resolution of remote sensing images via saliency-aware dynamic routing strategy},
  author={Wu, Hanlin and Ni, Ning and Zhang, Libao},
  journal={IEEE Transactions on Geoscience and Remote Sensing},
  volume={61},
  pages={1--17},
  year={2023},
  publisher={IEEE}
}

@inproceedings{remote_sensing_IR3,
  title={Effective cloud removal for remote sensing images by an improved mean-reverting denoising model with elucidated design space},
  author={Liu, Yi and Li, Wengen and Guan, Jihong and Zhou, Shuigeng and Zhang, Yichao},
  booktitle={Proceedings of the Computer Vision and Pattern Recognition Conference},
  pages={17851--17861},
  year={2025}
}

@inproceedings{contrastive_learning,
  title={A simple framework for contrastive learning of visual representations},
  author={Chen, Ting and Kornblith, Simon and Norouzi, Mohammad and Hinton, Geoffrey},
  booktitle={International conference on machine learning},
  pages={1597--1607},
  year={2020},
  organization={PmLR}
}

@inproceedings{MOCO,
  title={Momentum contrast for unsupervised visual representation learning},
  author={He, Kaiming and Fan, Haoqi and Wu, Yuxin and Xie, Saining and Girshick, Ross},
  booktitle={Proceedings of the IEEE/CVF conference on computer vision and pattern recognition},
  pages={9729--9738},
  year={2020}
}

@inproceedings{MPerceiver,
  title={Multimodal prompt perceiver: Empower adaptiveness generalizability and fidelity for all-in-one image restoration},
  author={Ai, Yuang and Huang, Huaibo and Zhou, Xiaoqiang and Wang, Jiexiang and He, Ran},
  booktitle={Proceedings of the IEEE/CVF Conference on Computer Vision and Pattern Recognition},
  pages={25432--25444},
  year={2024}
}

@article{AdaIR2,
  title={AdaIR: Exploiting underlying similarities of image restoration tasks with adapters},
  author={Chen, Hao-Wei and Xu, Yu-Syuan and Chan, Kelvin CK and Kuo, Hsien-Kai and Lee, Chun-Yi and Yang, Ming-Hsuan},
  journal={arXiv preprint arXiv:2404.11475},
  year={2024}
}

@inproceedings{Uformer,
  title={Uformer: A general u-shaped transformer for image restoration},
  author={Wang, Zhendong and Cun, Xiaodong and Bao, Jianmin and Zhou, Wengang and Liu, Jianzhuang and Li, Houqiang},
  booktitle={Proceedings of the IEEE/CVF conference on computer vision and pattern recognition},
  pages={17683--17693},
  year={2022}
}

@inproceedings{SRCNN,
  title={Learning a deep convolutional network for image super-resolution},
  author={Dong, Chao and Loy, Chen Change and He, Kaiming and Tang, Xiaoou},
  booktitle={European conference on computer vision},
  pages={184--199},
  year={2014},
  organization={Springer}
}

@inproceedings{deblurring1,
  title={A neural approach to blind motion deblurring},
  author={Chakrabarti, Ayan},
  booktitle={European conference on computer vision},
  pages={221--235},
  year={2016},
  organization={Springer}
}

@article{prompt_nlp1,
  title={Autoprompt: Eliciting knowledge from language models with automatically generated prompts},
  author={Shin, Taylor and Razeghi, Yasaman and Logan IV, Robert L and Wallace, Eric and Singh, Sameer},
  journal={arXiv preprint arXiv:2010.15980},
  year={2020}
}

@article{prompt_nlp2,
  title={Language models are few-shot learners},
  author={Brown, Tom and Mann, Benjamin and Ryder, Nick and Subbiah, Melanie and Kaplan, Jared D and Dhariwal, Prafulla and Neelakantan, Arvind and Shyam, Pranav and Sastry, Girish and Askell, Amanda and others},
  journal={Advances in neural information processing systems},
  volume={33},
  pages={1877--1901},
  year={2020}
}

@article{prompt_mutimodal1,
  title={Learning to prompt for vision-language models},
  author={Zhou, Kaiyang and Yang, Jingkang and Loy, Chen Change and Liu, Ziwei},
  journal={International Journal of Computer Vision},
  volume={130},
  number={9},
  pages={2337--2348},
  year={2022},
  publisher={Springer}
}

@inproceedings{prompt_mutimodal2,
  title={Visual-language prompt tuning with knowledge-guided context optimization},
  author={Yao, Hantao and Zhang, Rui and Xu, Changsheng},
  booktitle={Proceedings of the IEEE/CVF conference on computer vision and pattern recognition},
  pages={6757--6767},
  year={2023}
}

@inproceedings{dualprompt,
  title={Dualprompt: Complementary prompting for rehearsal-free continual learning},
  author={Wang, Zifeng and Zhang, Zizhao and Ebrahimi, Sayna and Sun, Ruoxi and Zhang, Han and Lee, Chen-Yu and Ren, Xiaoqi and Su, Guolong and Perot, Vincent and Dy, Jennifer and others},
  booktitle={European conference on computer vision},
  pages={631--648},
  year={2022},
  organization={Springer}
}

@article{remote_sensing_dehazing,
  title={A spatial--spectral adaptive haze removal method for visible remote sensing images},
  author={Shen, Huanfeng and Zhang, Chi and Li, Huifang and Yuan, Quan and Zhang, Liangpei},
  journal={IEEE Transactions on Geoscience and Remote Sensing},
  volume={58},
  number={9},
  pages={6168--6180},
  year={2020},
  publisher={IEEE}
}

@inproceedings{UNet,
  title={U-net: Convolutional networks for biomedical image segmentation},
  author={Ronneberger, Olaf and Fischer, Philipp and Brox, Thomas},
  booktitle={International Conference on Medical image computing and computer-assisted intervention},
  pages={234--241},
  year={2015},
  organization={Springer}
}

@inproceedings{MaIR,
  title={Mair: A locality-and continuity-preserving mamba for image restoration},
  author={Li, Boyun and Zhao, Haiyu and Wang, Wenxin and Hu, Peng and Gou, Yuanbiao and Peng, Xi},
  booktitle={Proceedings of the Computer Vision and Pattern Recognition Conference},
  pages={7491--7501},
  year={2025}
}

@article{yunjin_chen_denoising,
  title={Beyond a gaussian denoiser: Residual learning of deep cnn for image denoising},
  author={Zhang, Kai and Zuo, Wangmeng and Chen, Yunjin and Meng, Deyu and Zhang, Lei},
  journal={IEEE transactions on image processing},
  volume={26},
  number={7},
  pages={3142--3155},
  year={2017},
  publisher={IEEE}
}

@article{RICE,
  title={A remote sensing image dataset for cloud removal},
  author={Lin, Daoyu and Xu, Guangluan and Wang, Xiaoke and Wang, Yang and Sun, Xian and Fu, Kun},
  journal={arXiv preprint arXiv:1901.00600},
  year={2019}
}

@article{SSIM,
  title={Image quality assessment: from error visibility to structural similarity},
  author={Wang, Zhou and Bovik, Alan C and Sheikh, Hamid R and Simoncelli, Eero P},
  journal={IEEE transactions on image processing},
  volume={13},
  number={4},
  pages={600--612},
  year={2004},
  publisher={IEEE}
}

@article{prior_allinone,
  title={Training-Free Large Model Priors for Multiple-in-One Image Restoration},
  author={He, Xuanhua and Li, Lang and Wang, Yingying and Zheng, Hui and Cao, Ke and Yan, Keyu and Li, Rui and Xie, Chengjun and Zhang, Jie and Zhou, Man},
  journal={arXiv preprint arXiv:2407.13181},
  year={2024}
}

@article{UniECS,
  title={UniECS: Unified Multimodal E-Commerce Search Framework with Gated Cross-modal Fusion},
  author={Liang, Zihan and Ma, Yufei and Qian, ZhiPeng and Dai, Huangyu and Wang, Zihan and Chen, Ben and Lei, Chenyi and Ding, Yuqing and Li, Han},
  journal={arXiv preprint arXiv:2508.13843},
  year={2025}
}

@inproceedings{AIGC_SR,
  title={Evaluating and improving compositional text-to-visual generation},
  author={Li, Baiqi and Lin, Zhiqiu and Pathak, Deepak and Li, Jiayao and Fei, Yixin and Wu, Kewen and Xia, Xide and Zhang, Pengchuan and Neubig, Graham and Ramanan, Deva},
  booktitle={Proceedings of the IEEE/CVF Conference on Computer Vision and Pattern Recognition},
  pages={5290--5301},
  year={2024}
}

@article{Manga109,
  title={Sketch-based manga retrieval using manga109 dataset},
  author={Matsui, Yusuke and Ito, Kota and Aramaki, Yuji and Fujimoto, Azuma and Ogawa, Toru and Yamasaki, Toshihiko and Aizawa, Kiyoharu},
  journal={Multimedia tools and applications},
  volume={76},
  pages={21811--21838},
  year={2017},
  publisher={Springer}
}

@article{snow100k,
  title={Desnownet: Context-aware deep network for snow removal},
  author={Liu, Yun-Fu and Jaw, Da-Wei and Huang, Shih-Chia and Hwang, Jenq-Neng},
  journal={IEEE Transactions on Image Processing},
  volume={27},
  number={6},
  pages={3064--3073},
  year={2018},
  publisher={IEEE}
}

@inproceedings{SHIQ,
  title={A multi-task network for joint specular highlight detection and removal},
  author={Fu, Gang and Zhang, Qing and Zhu, Lei and Li, Ping and Xiao, Chunxia},
  booktitle={Proceedings of the IEEE/CVF Conference on Computer Vision and Pattern Recognition},
  pages={7752--7761},
  year={2021}
}

@article{lin_tpami,
  title={Re-boosting self-collaboration parallel prompt gan for unsupervised image restoration},
  author={Lin, Xin and Zhou, Yuyan and Yue, Jingtong and Ren, Chao and Chan, Kelvin CK and Qi, Lu and Yang, Ming-Hsuan},
  journal={IEEE Transactions on Pattern Analysis and Machine Intelligence},
  year={2025},
  publisher={IEEE}
}

@article{4kagent,
  title={4kagent: agentic any image to 4k super-resolution},
  author={Zuo, Yushen and Zheng, Qi and Wu, Mingyang and Jiang, Xinrui and Li, Renjie and Wang, Jian and Zhang, Yide and Mai, Gengchen and Wang, Lihong V and Zou, James and others},
  journal={arXiv preprint arXiv:2507.07105},
  year={2025}
}

@inproceedings{MUSIQ,
  title={Musiq: Multi-scale image quality transformer},
  author={Ke, Junjie and Wang, Qifei and Wang, Yilin and Milanfar, Peyman and Yang, Feng},
  booktitle={Proceedings of the IEEE/CVF international conference on computer vision},
  pages={5148--5157},
  year={2021}
}

@inproceedings{CLIPIQA,
  title={Exploring clip for assessing the look and feel of images},
  author={Wang, Jianyi and Chan, Kelvin CK and Loy, Chen Change},
  booktitle={Proceedings of the AAAI Conference on Artificial Intelligence},
  volume={37},
  pages={2555--2563},
  year={2023}
}

@article{NIQE,
  title={A feature-enriched completely blind image quality evaluator},
  author={Zhang, Lin and Zhang, Lei and Bovik, Alan C},
  journal={IEEE Transactions on Image Processing},
  volume={24},
  number={8},
  pages={2579--2591},
  year={2015},
  publisher={IEEE}
}

@article{FID,
  title={Gans trained by a two time-scale update rule converge to a local nash equilibrium},
  author={Heusel, Martin and Ramsauer, Hubert and Unterthiner, Thomas and Nessler, Bernhard and Hochreiter, Sepp},
  journal={Advances in neural information processing systems},
  volume={30},
  year={2017}
}

@inproceedings{MANIQA,
  title={Maniqa: Multi-dimension attention network for no-reference image quality assessment},
  author={Yang, Sidi and Wu, Tianhe and Shi, Shuwei and Lao, Shanshan and Gong, Yuan and Cao, Mingdeng and Wang, Jiahao and Yang, Yujiu},
  booktitle={Proceedings of the IEEE/CVF conference on computer vision and pattern recognition},
  pages={1191--1200},
  year={2022}
}

@article{Qwen3,
  title={Qwen3 technical report},
  author={Yang, An and Li, Anfeng and Yang, Baosong and Zhang, Beichen and Hui, Binyuan and Zheng, Bo and Yu, Bowen and Gao, Chang and Huang, Chengen and Lv, Chenxu and others},
  journal={arXiv preprint arXiv:2505.09388},
  year={2025}
}

@article{Qwen2.5-vl,
  title={Qwen2. 5-vl technical report},
  author={Bai, Shuai and Chen, Keqin and Liu, Xuejing and Wang, Jialin and Ge, Wenbin and Song, Sibo and Dang, Kai and Wang, Peng and Wang, Shijie and Tang, Jun and others},
  journal={arXiv preprint arXiv:2502.13923},
  year={2025}
}

@InProceedings{ntire2025,
    author={Li, Xin and Jin, Yeying and Jin, Xin and Wu, Zongwei and Li, Bingchen and Wang, Yufei and Yang, Wenhan and Li, Yu and Chen, Zhibo and Wen, Bihan and others},
    title     = {NTIRE 2025 Challenge on Day and Night Raindrop Removal for Dual-Focused Images: Methods and Results},
    booktitle = {Proceedings of the IEEE/CVF Conference on Computer Vision and Pattern Recognition (CVPR) Workshops},
    month     = {June},
    year      = {2025},
    pages     = {1172-1183}
}
\bibliographystyle{iclr2026_conference}
\newpage
\appendix
\section*{Appendix}
\section{Use of Large Language Models}
Large language models were used solely for light editing tasks including grammar correction, spelling checks, and minor phrasing improvements to enhance clarity and concision.

\section{Discussion on Broader Implication and Limitation}
In this paper, we propose a dual-prompt-pool approach tailored to MD-AiOIR. In particular, our analysis and experiments reveal that different tasks/domains share some common knowledge, while still preserving specific differences. Modeling these `shared + specific' priors can significantly reduce the learning difficulty when facing more tasks and domains, and help improve restoration performance. We believe this insight opens a new perspective for moving image restoration toward more unified and scalable models across multiple domains. There are also several limitations in our framework. The training cost of our framework is a little bit higher than that of single-domain and single-task models. In addition, although our method demonstrates strong zero-shot behavior, extending it to a broader range of domains remains an interesting direction. It is worth further exploring how to model the shared and different prior knowledge between various restoration tasks more efficiently and interpretably.

\section{Detailed Experimental Setting}
\label{details}
\textbf{Datasets.} The training datasets for each task are as follows: Natural image SR is trained on the DF2K~\citep{DIV2K,F2K} dataset (DIV2K + Flickr2K) with 4× bicubic downsampling. Natural image deraining is trained using Rain100L~\citep{Rain100L}. Natural image deblurring uses the GoPro~\citep{GoPro} dataset. Following~\citet{amir,Restore-rwkv-allinone-medical}, medical MRI SR is trained on the IXI\footnote{https://brain-development.org/ixi-dataset/} MRI dataset. Medical CT denoising uses dataset from the 2016 NIH AAPM-Mayo Clinic Low-Dose CT Grand Challenge~\citep{AAPM-CT-DN}. Medical PET synthesis is trained on the PolarStar m660 dataset, where both low-quality (LQ) and high-quality (HQ) PET images are reconstructed via the standard OSEM~\citep{osem-pet} method. Remote sensing image SR is trained on the UCMerced Land Use~\citep{UCMerced} dataset with 4× bicubic downsampling. Remote sensing cloud removal is trained using CUHK CR1~\citep{cuhk_cr1} dataset. Remote sensing dehazing is trained on RICE1~\citep{RICE}, which provides hazy and clean image pairs. Table \ref{table.datasets} presents the detailed numbers of training and testing images for each dataset. Data augmentation including random cropping, horizontal flipping, and rotation are applied to improve robustness. We unify all inputs by converting every dataset sample into a standard 3-channel visual image before feeding it into the network. Natural and remote-sensing datasets are already stored as RGB images. For medical datasets that are not originally stored in RGB format, we first convert them into grayscale images, and then replicate the single channel three times to obtain a 3-channel representation. All images are stored in a visual format (PNG or JPG), ensuring consistent dimensionality.

\begin{table}[h]
  \scriptsize
  \centering
  % \large
  \caption{Detailed description of the datasets utilized.}
  \resizebox{\textwidth}{!}{
  \label{table.datasets}
  \centering
  \begin{tabular}{c|ccccccc}
    \toprule
    Datasets & DF2K & Rain100L & GoPro & IXI MRI & AAPM-Mayo & PolarStar m660 & UCMerced \\
    % \cmidrule(r){1-2}
    \midrule
    Train & 3450 & 200 & 2103 & 40500 & 18351 & 27837 & 1800  \\
    Test & 100 & 100 & 1111 & 11400 & 211 & 2044 & 300\\
    Tasks & Natural SR & Natural Deraining  & Natural Deblurring & MRI SR & CT Denoising & PET Synthesis & RSI SR\\
    \bottomrule
    \toprule
    Datasets & CUHK CR1 & RICE1 \\
    % \cmidrule(r){1-2}
    \midrule
    Train   & 534 & 400 \\
    Test   & 134 & 100   \\
    Tasks   & Cloud Removal & RSI Dehazing  \\
    \bottomrule
  \end{tabular}
  }
\end{table}

\textbf{Implementation Details.} We train our model under PyTorch~\citep{pytorch} framework using the Adam~\citep{adam} optimizer with $\beta_{1} = 0.9$, $\beta_{2} = 0.99$. The learning rate is initialized at $4 \times 10^{-4}$ with cosine annealing. Batch size is set to 12, and we train for 1000K iterations on NVIDIA RTX 5090 GPUs. We set the diversity threshold $\tau_{\text{div}} = 0.1$. The loss weights are set to $\lambda_{\text{align}} = 1.0$, $\lambda_{\text{div}} = 0.1$, $\lambda_{\text{con}} = 0.1$, and $\lambda_{\text{bal}} = 0.1$. For each prompt, the key is defined as a 1×1024 vector, while the value is set to 2×1024. The numbers of prompts in both the task and domain prompt pools are set to 15, with top-k selection configured as k=3 for the task prompt pool and k=5 for the domain prompt pool. The projector is a 3-layer lightweight CNN (mainly including Conv2d, AdaptiveAvgPool2d, and MLP), and the dimensionality of its output is 1024. The used LLaVA-v1.5-7B is primarily built from CLIP’s ViT-L/14 visual encoder~\citep{CLIP} and Vicuna-7B~\citep{vicuna}, a language model based on the LLaMA~\citep{LLaMA} architecture. We instruct the LLaVA-v1.5-7B to produce concise descriptions, and a very small number of texts that exceed the CLIP text encoder length limit will be automatically truncated to 77 tokens. It is worth noting that the LLaVA-v1.5-7B and the CLIP text encoder are only used during the training phase. During inference, neither the LLaVA-v1.5-7B nor CLIP is required, thereby introducing no additional inference overhead. In Table \ref{table: Examples of LLaVA-generated textual descriptions}, we present several examples of LLaVA-generated textual descriptions for medical, remote sensing, and natural images. These descriptions provide concise and discriminative domain cues that are sufficient for alignment. 

\textbf{Contrastive Regularization.} In addition to the diversity regularization and prompt entropy regularization described in the main text, we also introduce a contrastive regularization to enhance the sensitivity of instance-level prompt selection. Specifically, we adopt a contrastive objective to align the query with the keys of their most relevant prompts while pushing them away from the keys of unrelated prompts. The contrastive regularization is defined as:
\begin{equation}
    \mathcal{L}_{\text{con}} 
    = - \log \frac{\exp(\langle \mathbf{q}, \mathbf{k}^+\rangle / \tau)}
    {\exp(\langle \mathbf{q}, \mathbf{k}^+\rangle / \tau) + 
    \sum_{\mathbf{k}^-} \exp(\langle \mathbf{q}, \mathbf{k}^-\rangle / \tau)},
\end{equation}
where $\mathbf{q}$ denotes the query and $\{\mathbf{k}^+\}$ and $\{\mathbf{k}^-\}$ denote the sets of positive (selected) and negative (non-selected) prompt keys.

\begin{table*}[t]
\centering
 \Large
\caption{Examples of LLaVA-generated textual descriptions for medical, remote sensing, and natural images.}
\resizebox{\textwidth}{!}{
\centering
\begin{tabular}{c}
\toprule
\textbf{Medical Images}\\
\midrule
“The image is a black and white photo of a brain, with a close up view of the frontal lobe.” \\
“The image is a black and white picture of a human body, specifically focusing on the abdominal area.” \\
“The image is a grayscale medical image with low overall contrast, featuring a bright background and a few small dark foci near the center.”\\
\midrule
\textbf{Remote Sensing Images}\\
\midrule
 “The image is a large, green field with a white snowy ground, and it is a winter scene.”\\
“The image is a bird's-eye view of a forest with low brightness, where a river flows through it and a highway is visible nearby.”\\
“The image is a bird's eye view of a large building with a pool, surrounded by palm trees.”\\
\midrule
\textbf{Natural Images} \\
\midrule
“The image features two women in red outfits walking through a grassy field, one of them carrying a basket.” \\
“The image depicts a beautiful garden entrance featuring a red door with a large vine growing up the side of the building.” \\
“A colorful image taken from a boat shows tourists on red seats under a canopy looking over vivid blue sea toward a detailed coastal cityscape.”\\
\bottomrule

\end{tabular}
}
\label{table: Examples of LLaVA-generated textual descriptions}
\end{table*}

\textbf{Compared Methods.} We compare our DATPRL-IR with several SOTA AiOIR methods~\citep{moceir, adair, tian, amir, promptir, transweather} and classic image restoration baselines~\citep{NAFNet, restormer, SWINIR, MPRNet}. To ensure a fair comparison, all compared methods are retrained from scratch on the same datasets as ours using their original loss functions and training strategies as specified in their papers. All remaining training conditions were kept identical to ours, including the train/test splits, the domain-balanced sampling strategy, data augmentations, and other shared hyperparameters.

\begin{table*}
    \centering
    \Large
    \caption{Quantitative comparison between our method and other SOTA methods on 3 domains \& 9 tasks experimental setting. The best and second-best metrics are highlighted in \textbf{bold} and \underline{underline}.} 
     \resizebox{\textwidth}{!}{
    \centering
    \begin{tabular}{c|cccccccccccccccccccc}
    \toprule
     \multicolumn{1}{c|}{\multirow{1}{*}{Image Domain}} & \multicolumn{3}{c}{Natural Image} & \multicolumn{3}{c}{Medical Image} & \multicolumn{3}{c}{Remote Sensing Image} \\
    \midrule
     \multicolumn{1}{c|}{\multirow{2}{*}{\makecell{Task \& \\ Dataset}}} & \multicolumn{1}{c}{SR} & \multicolumn{1}{c}{Deraining}& \multicolumn{1}{c}{Deblurring} & \multicolumn{1}{c}{MRI SR} & \multicolumn{1}{c}{CT Denoising}& \multicolumn{1}{c}{PET Synthesis} &\multicolumn{1}{c}{RSI SR} & \multicolumn{1}{c}{Cloud Removal}&\multicolumn{1}{c}{RSI Dehazing}  \\
      & \multicolumn{1}{c}{on DIV2K-Val} & \multicolumn{1}{c}{on Rain100L}& \multicolumn{1}{c}{on GoPro} & \multicolumn{1}{c}{on IXI MRI} & \multicolumn{1}{c}{on AAPM-Mayo} & \multicolumn{1}{c}{on PolarStar m660} & \multicolumn{1}{c}{on UCMerced} & \multicolumn{1}{c}{on CUHK CR1} & \multicolumn{1}{c}{on RICE1}  \\
    \midrule
    \multicolumn{5}{l}{Single-Task Method}  & \\
    \midrule
    MPRNet& 28.32 / 0.8067 & 37.55 / 0.9797 & 28.02 /  0.8570 & 26.69 / 0.8871 & 33.54 / 0.9253 & 36.72 / 0.9475 & 27.47 /0.7646 & 25.20 / 0.7334 & 25.66 / 0.9268\\
     SwinIR & 28.80 / 0.8109 & 37.55 / 0.9802 & 28.17 / 0.8510 & 26.49 / 0.8844 & 33.63 /0.9251 & 36.78 / 0.9468 & 27.52 / 0.7670 & 25.39 / 0.7249 & 25.42 / 0.9244 \\
    Restormer& 28.63 / 0.8150 & \underline{38.45} / 0.9833 & 29.06 / 0.8805 & 27.43 / 0.8992 & 33.70 / 0.9269 & \textbf{37.20} / \textbf{0.9509} & 27.94 / 0.7827 & 25.66 / 0.7451 & 26.07 / 0.9286\\
    NAFNet  & 28.64 / 0.8128 & 37.31 / 0.9783 & 29.20 / 0.8828 & 27.31 / 0.8977 & 33.66 / 0.9267 & 37.14 / 0.9505 & 27.80 / 0.7774 & \underline{25.96} / \underline{0.7591} & 26.45 / 0.9215	 \\
    \midrule
        \multicolumn{5}{l}{All-in-One Method}  & \\
    \midrule
     Transweather & 28.16 / 0.7951 & 32.41 / 0.9392 & 26.53 / 0.8116 & 25.55 / 0.8638 & 32.98 / 0.9209 & 36.47 / 0.9440 & 26.90 / 0.7402 & 25.35 / 0.7193 & 25.18 / 0.9226\\
      PromptIR & \underline{28.86} / 0.8127 & 38.21 / 0.9811 & 28.79 / 0.8749 & 27.31 / 0.8964 & 33.66 / 0.9265 & 37.03 / 0.9495 & 27.91 / 0.7809  & 25.33 / 0.7360 & \underline{26.86} / \textbf{0.9399} \\
       AdaIR& 28.24 / \underline{0.8153}& 38.40 / \underline{0.9834} & 29.21 / 0.8835 & 27.52 / \underline{0.9013} & 33.70 / 0.9270 & \underline{37.17} / \underline{0.9508} & 27.96 / \underline{0.7837} & 25.87 / 0.7528 & 25.40 / 0.9274 \\
        MoCEIR& 28.68 / 0.8152 & 38.26 / 0.9827 & \underline{29.32} / \underline{0.8855} & \underline{27.62} / 0.9003  & \underline{33.72} / \textbf{0.9278 }& 37.16 / 0.9502 & \underline{ 28.00} / 0.7809 & 25.83 / 0.7586 & 26.31 / \underline{0.9395 }\\
    \midrule
            \multicolumn{5}{l}{Muti-Domain All-in-One Method}  & \\
    \midrule
     \rowcolor{yellow!8}
     DATPRL-IR-\textbf{\textcolor{red}{9T}} (Ours)& \textbf{29.05} / \textbf{0.8181} & \textbf{39.67} / \textbf{0.9867} & \textbf{29.57} / \textbf{0.8881} & \textbf{27.86} / \textbf{0.9045} & \textbf{33.77} / \underline{0.9273} & 37.12 / 0.9502 & \textbf{28.31} / \textbf{0.7913} & \textbf{26.00} / \textbf{0.7592} & \textbf{26.94} / 0.9347 \\
    \bottomrule
    \end{tabular}
    }
\label{table: 3 domains 9 tasks}
\end{table*}

\section{More Experiment Results}
\label{additional_results}
\subsection{Multi-domain all-in-one image restoration}
\begin{figure}[t]
  \centering
  \setlength{\abovecaptionskip}{0.08cm}
  \includegraphics[width=1.0\textwidth]{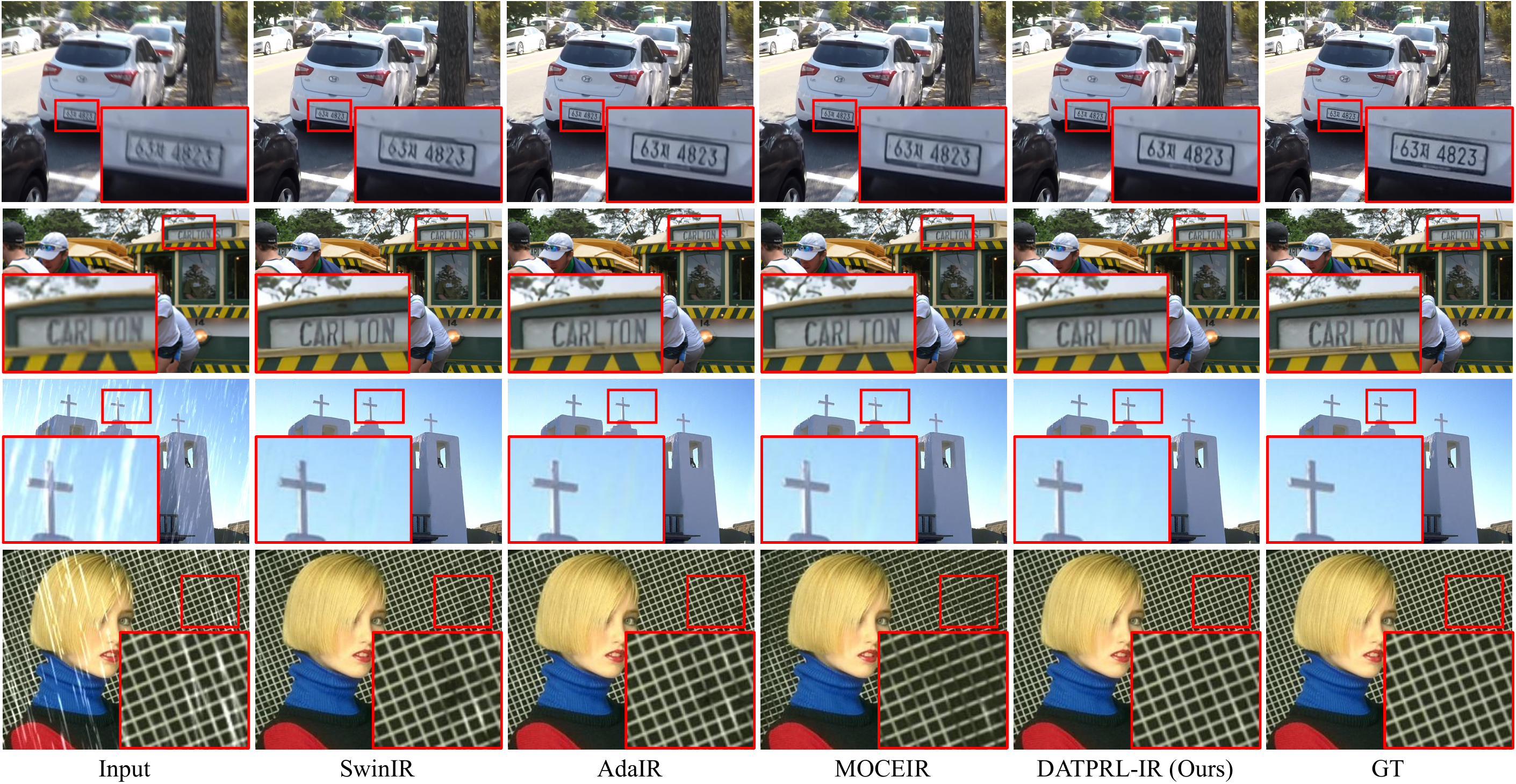}
  \caption{Comparison of our DATPRL-IR with other SOTA methods on Natural Images.}
  \label{fig.Visual_Compare3}
\end{figure}

\begin{figure}[t]
  \centering
  \setlength{\abovecaptionskip}{0.08cm}
  \includegraphics[width=1.0\textwidth]{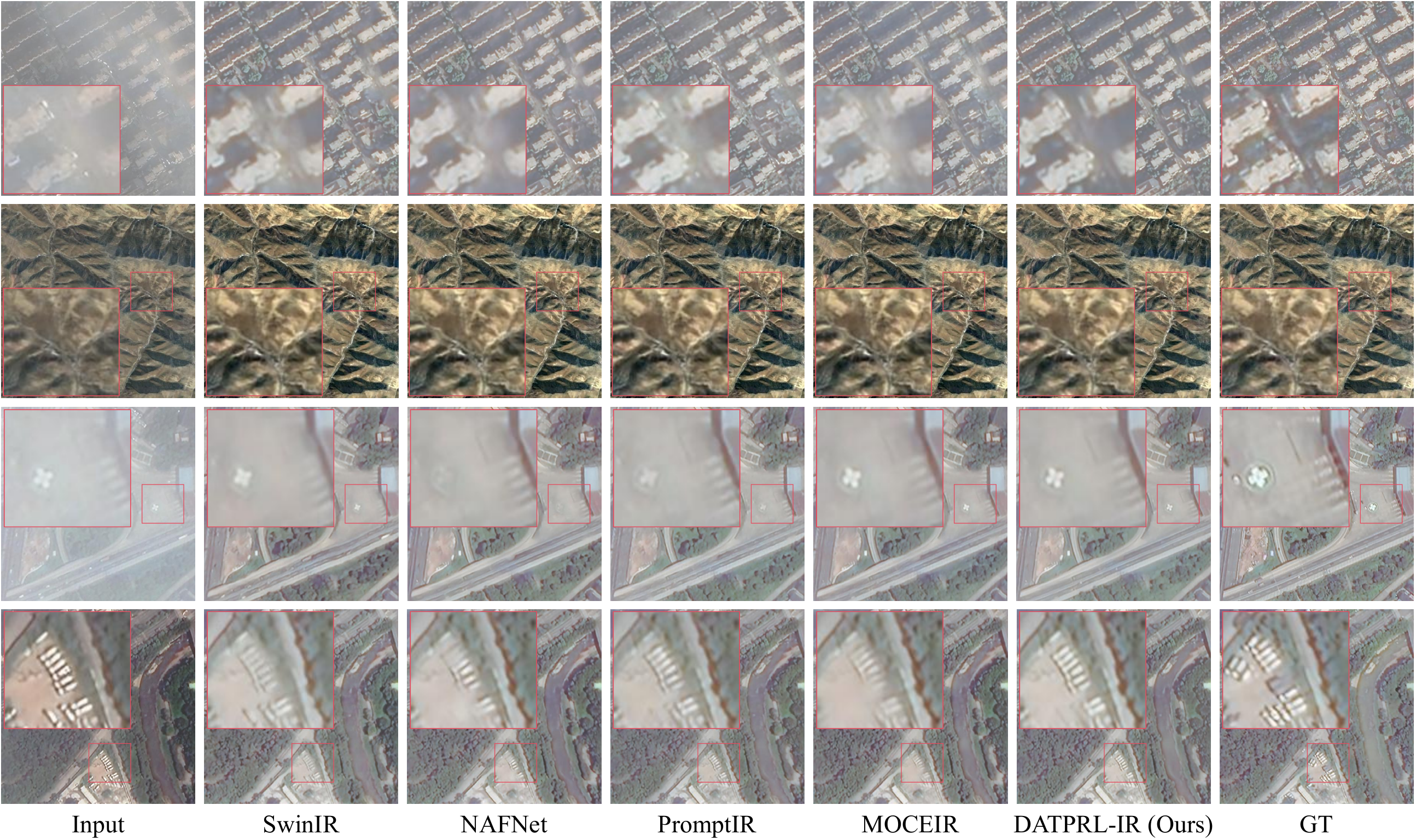}
  \caption{Comparison of our DATPRL-IR with other SOTA methods on Remote Sensing Images.}
  \label{fig.Visual_Compare4}
\end{figure}

\begin{figure}[t]
  \centering
  \setlength{\abovecaptionskip}{0.08cm}
  \includegraphics[width=1.0\textwidth]{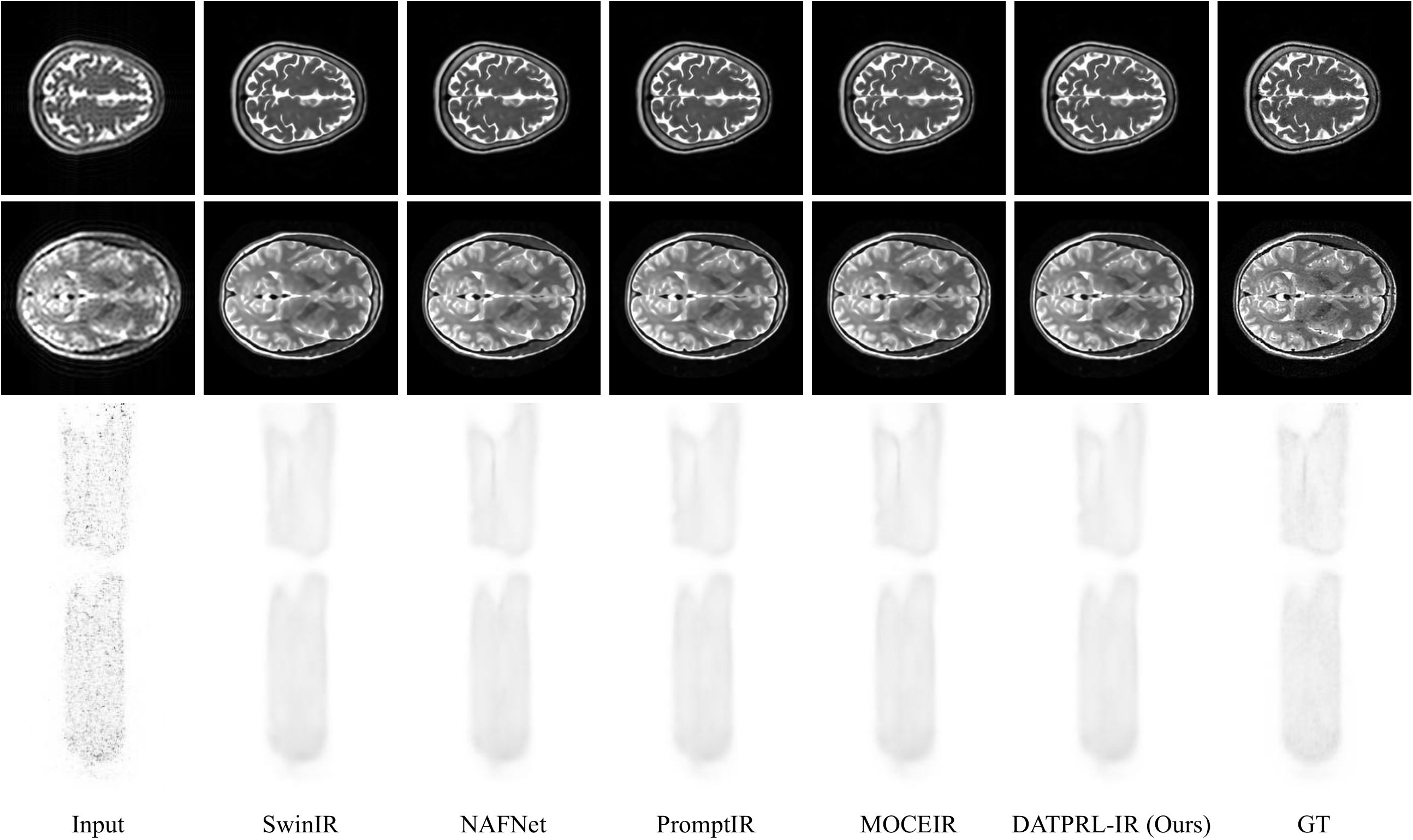}
  \caption{Comparison of our DATPRL-IR with other SOTA methods on Medical Images.}
  \label{fig.Visual_Compare5}
\end{figure}

We present detailed quantitative comparisons between our method and other approaches in Table \ref{table: 3 domains 9 tasks}. Our DATPRL-IR surpasses other SOTA methods across most tasks, demonstrating the superiority of our proposed domain-aware task prompt representation learning. Moreover, compared with the 6-task setting, it exhibits no performance drop on the original six tasks, but even achieves further improvements. We also provide additional qualitative visual comparisons in Figure \ref{fig.Visual_Compare3}, Figure \ref{fig.Visual_Compare4}, and Figure \ref{fig.Visual_Compare5}. Obviously, compared to other methods, our method is able to remove degradations clearer and reconstruct more image details. These results significant demonstrate the effectiveness of our method.

\subsection{Zero-shot and generalization performance}
To demonstrate the zero-shot and generalization capability of our model, we add another three evaluation settings: (1) Zero-shot on unseen image domains; 
(2) Generalization to unseen distributions; (3) Zero-shot on completely unseen restoration tasks. All experiments are performed using the well-trained model, and no fine-tuning is applied to any method.

\begin{figure}[t]
  \centering
  \setlength{\abovecaptionskip}{0.08cm}
  \includegraphics[width=1.0\textwidth]{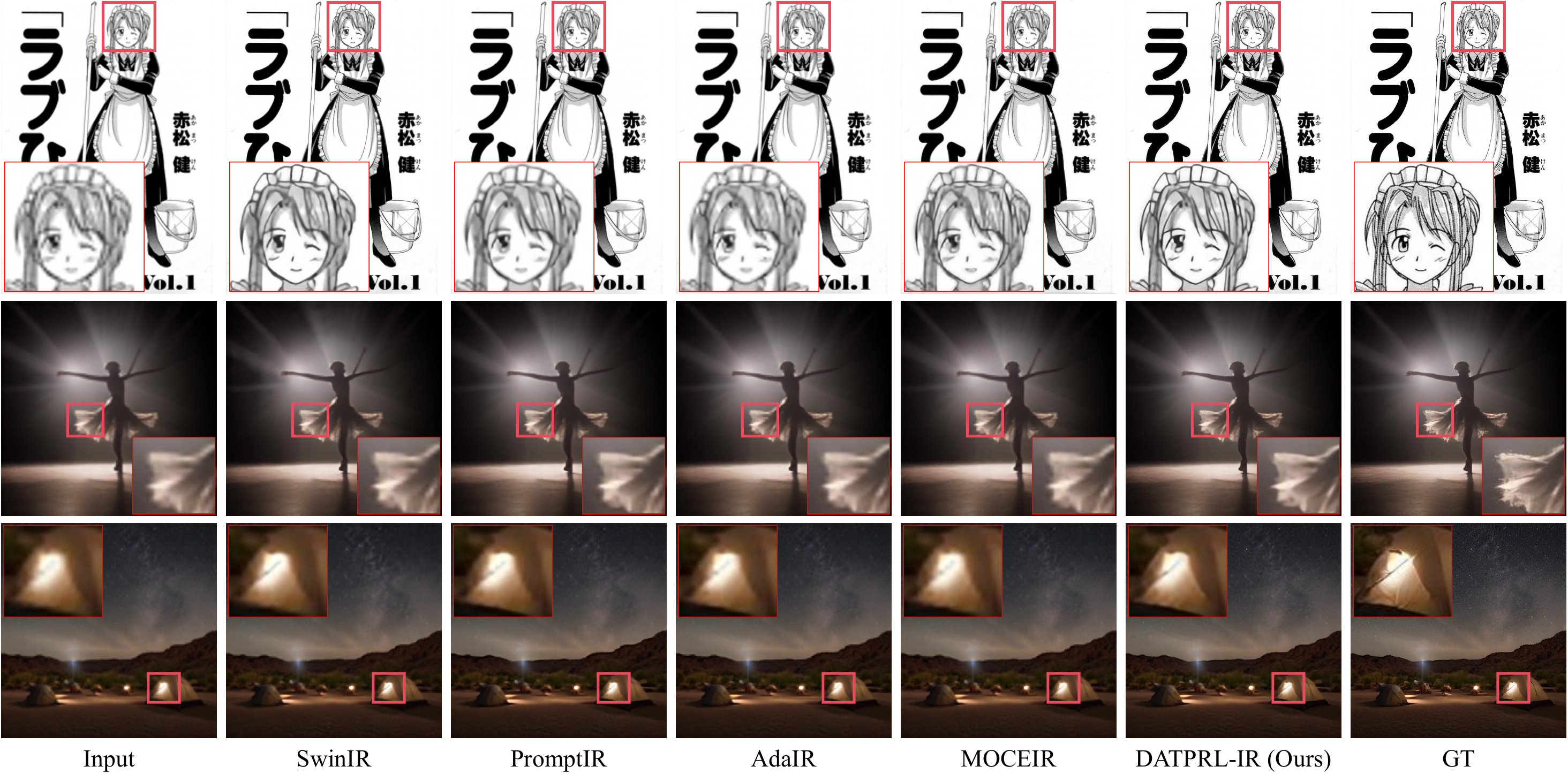}
  \caption{Zero-shot comparison of our DATPRL-IR with other SOTA methods on AIGC and comic domains. Please zoom in for better visualization.}
  \label{fig.AIGC_Manga109_compare}
\end{figure}

\textbf{Zero-shot on unseen image domains.} We test zero-shot super-resolution (SR) performance on two image domains never seen during training: AI-generated content (AIGC) images and comic images. Following~\citep{4kagent}, we use the standard AIGC benchmark GenAI-Bench~\citep{AIGC_SR} for testing. Specifically, we randomly selected 100 images from each of the three subsets SDXL-Turbo, DeepFloyd, and Midjourney, and performed 4x bicubic downsampling to construct our AIGC SR test set. For comic image SR, we chose the most commonly used Manga109\citep{Manga109} as the test set. As shown in Table \ref{table: AIGC}, our method comprehensively outperforms other approaches in two unseen image domains, achieving PSNR improvements of 0.87dB and 0.45dB over the prompt-based PromptIR on SDXL-Turbo and Manga109, respectively. Furthermore, the visual comparison in Figure \ref{fig.AIGC_Manga109_compare} demonstrates that our method can reconstruct much clearer texture details, with fewer blurry edges and artifacts.

\begin{table*}
    \centering
    \Large
    \caption{Quantitative zero-shot super-resolution comparison between our method and other SOTA methods on AIGC and comic images. The best metrics are highlighted in \textbf{bold}.}  
    \resizebox{0.8\textwidth}{!}{
    \centering
    \begin{tabular}{cc|cccccc|cc}
    \toprule
     \multicolumn{2}{c|}{\multirow{1}{*}{Image Domain \& Task}} & \multicolumn{6}{c|}{AI-Generated Content (AIGC) SR} & \multicolumn{2}{c}{Comic Image SR} \\
    \cmidrule(lr){1-10}
     \multicolumn{2}{c|}{\multirow{1}{*}{Dataset}} & 
       \multicolumn{2}{c}{on SDXL-Turbo} & \multicolumn{2}{c}{on DeepFloyd} & \multicolumn{2}{c|}{on Midjourney} & \multicolumn{2}{c}{on Manga109}  \\
    \cmidrule(lr){1-10}
      \multicolumn{1}{c}{\multirow{1}{*}{Method}} &\multicolumn{1}{c}{\multirow{1}{*}{Year}} & PSNR $\uparrow$ & SSIM $\uparrow$ & PSNR $\uparrow$ & SSIM $\uparrow$ & PSNR $\uparrow$ & SSIM $\uparrow$ & PSNR $\uparrow$ & SSIM $\uparrow$   \\
    \midrule
    \multicolumn{2}{c}{Single-Task Method}  & \\
    \midrule
    MPRNet& CVPR2021 &  31.13 &  0.8730	& 26.76 & 0.8254 &  27.19 & 0.7656  & 27.13 &  0.8612\\
    SwinIR& ICCVW2021 &  31.20 &  0.8709 & 27.41 & 0.8344 & 27.19 & 0.7627 & 27.76 &  0.8697\\
    Restormer& CVPR2022 & 32.95 & 0.8985 & 28.81 & 0.8675 & 28.24 & 0.8005  & 28.13 & 0.8758\\
      NAFNet & ECCV2022  & 32.67 & 0.8959 &  28.59 & 0.8637 &  28.12 & 0.7965 &  27.98 & 0.8727\\
      \midrule
    \multicolumn{2}{c}{All-in-One Method}  & \\
    \midrule
     Transweather & CVPR2022 & 32.49 & 0.8916 &  27.73 & 0.8456 & 27.81 & 0.7866 &  26.00 & 0.8330\\
      PromptIR & NeurIPS2023  & 32.38 & 0.8933 &   28.56 & 0.8642 &  27.90 & 0.7916 &   28.28 & 0.8756\\
     AdaIR & ICLR2025 & 32.95 & 0.8989 &  28.87 & 0.8687 & 28.23 & 0.8008 & 28.05 & 0.8758\\
      MoCEIR & CVPR2025 & 33.00 & 0.8995   & 28.90 & 0.8680  & 28.30 & 0.8027 & 28.10 & 0.8745	 \\
      \midrule
        \multicolumn{5}{l}{Muti-Domain All-in-One Method}  & \\
    \midrule
    \rowcolor{yellow!8}
       \multicolumn{2}{c}{DATPRL-IR \ \  (Ours)} & \textbf{33.25} & \textbf{0.9017}& \textbf{29.07} & \textbf{0.8713} &  \textbf{28.38} & \textbf{0.8045} & \textbf{28.73} & \textbf{0.8828}\\
    \bottomrule
    \end{tabular}
    }
\label{table: AIGC}
\end{table*}

\begin{figure}[t]
  \centering
  \setlength{\abovecaptionskip}{0.08cm}
  \includegraphics[width=1.0\textwidth]{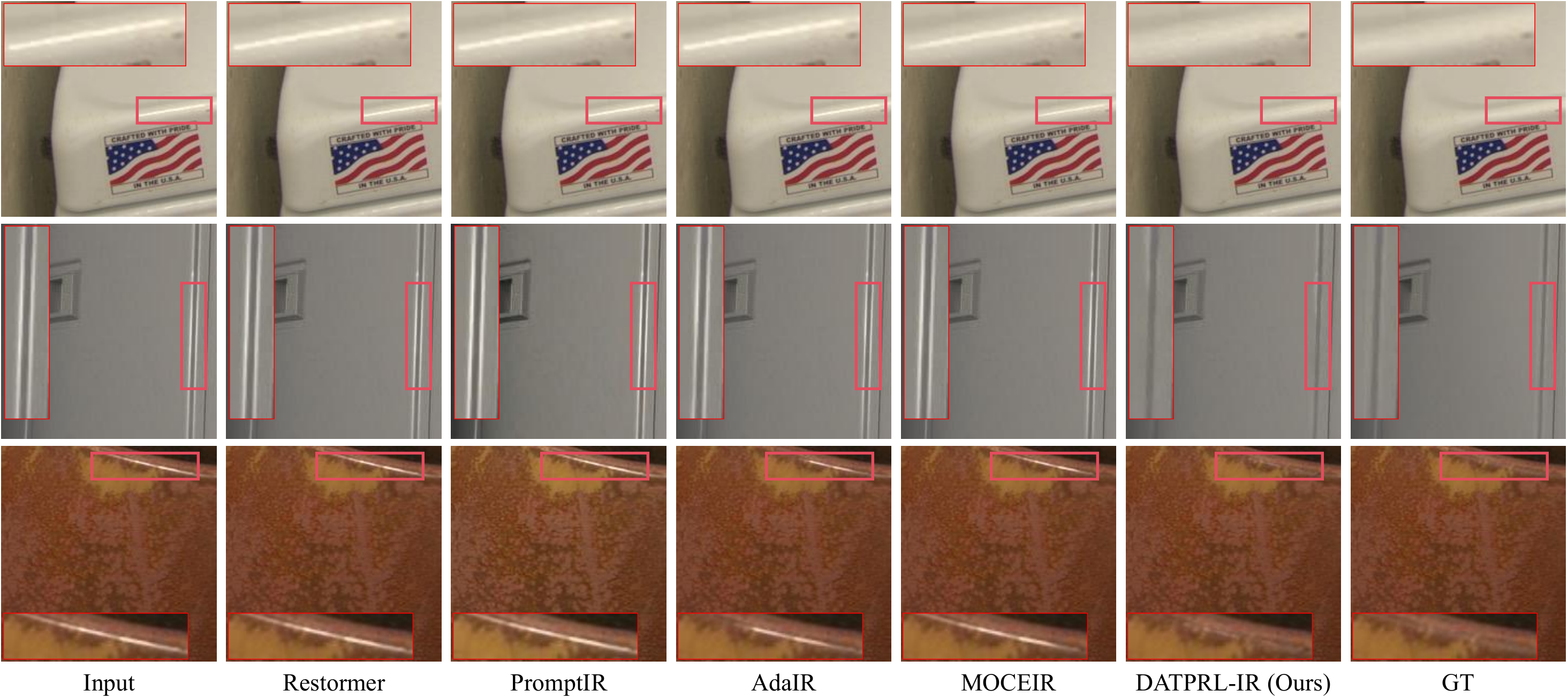}
  \caption{Zero-shot comparison of our DATPRL-IR with other SOTA methods on specular highlight removal task. Please zoom in for better visualization.}
  \label{fig.zeroshot_highlight_removal}
\end{figure}

\begin{figure}[t]
  \centering
  \setlength{\abovecaptionskip}{0.08cm}
  \includegraphics[width=1.0\textwidth]{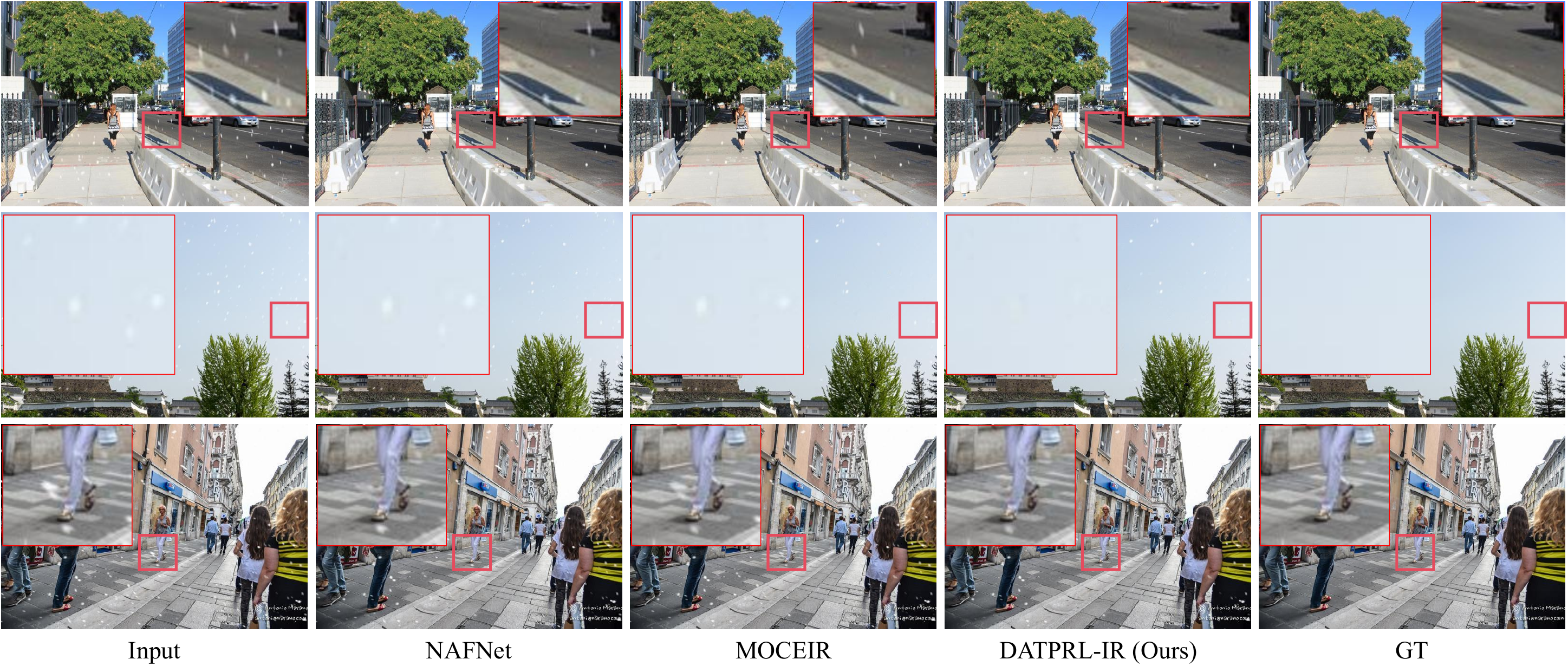}
  \caption{Zero-shot comparison of our DATPRL-IR with other SOTA methods on image desnowing task. Please zoom in for better visualization.}
  \label{fig.zeroshot_desnow}
\end{figure}

\textbf{Zero-shot on completely unseen restoration tasks.} We further test zero-shot performance on 2 tasks not included in training: natural image desnowing, and specular highlight removal. For image desnowing, following~\citep{lin_tpami}, we randomly select 2,000 pairs from the Snow100K-S~\citep{snow100k} test subset for evaluation. For specular highlight removal, we evaluate all methods on the large-scale real-world benchmark SHIQ~\citep{SHIQ}, which contains 1000 real image pairs for testing. As shown in Table \ref{table: image desnowing}, our method surpasses the SOTA MoCEIR by 0.92 dB and 0.55 dB in PSNR on specular highlight removal and image desnowing tasks respectively, without any fine-tuning. At the same time, Figure \ref{fig.zeroshot_highlight_removal} and Figure \ref{fig.zeroshot_desnow} show the comparison of the visual results of our method with other SOTA methods on these two completely unseen tasks. Our approach demonstrates extremely strong zero-shot capability, especially in specular highlight removal, where our method can accurately remove degradation and reconstruct lost textures, while other methods can hardly remove any highlights.

\begin{table*}
    \centering
    \Large
    % \captionsetup{skip=5pt}
    \caption{Quantitative zero-shot comparison between our method and other SOTA methods on unseen image specular highlight removal, and image desnowing. The best metrics are highlighted in \textbf{bold}.}
    \resizebox{0.58\textwidth}{!}{
    \centering
    \begin{tabular}{cc|cccc}
    \toprule
    \multicolumn{2}{c|}{\multirow{2}{*}{Task \& Dataset}} & \multicolumn{2}{c}{Highlight Removal} & \multicolumn{2}{c}{Image Desnowing}  \\
      & & \multicolumn{2}{c}{on SHIQ} & \multicolumn{2}{c}{on Snow100K-S} \\
    \cmidrule(lr){1-6}
      Method & Year & PSNR $\uparrow$ & SSIM $\uparrow$ & PSNR $\uparrow$ & SSIM $\uparrow$   \\
    \midrule
    \multicolumn{2}{c}{Single-Task Method}  & \\
    \midrule
    MPRNet& CVPR2021 & 25.83 & 0.9297 & 26.09 & 0.8882\\
    Restormer& CVPR2022 & 25.67 & 0.9295 & 26.57 & 0.8955\\
      NAFNet & ECCV2022  &  25.89 & 0.9298 & 26.27 & 0.8923 \\
      \midrule
    \multicolumn{2}{c}{All-in-One Method}  & \\
    \midrule
     Transweather & CVPR2022 & 26.20 & 0.9315 & 26.36 & 0.8870 \\

      PromptIR & NeurIPS2023  & 24.87 &  0.9239 &  26.42 & 0.8923\\

     AdaIR & ICLR2025 & 25.35 & 0.9263 & 26.54 & 0.8951\\

      MoCEIR & CVPR2025 & 25.30 & 0.9259 & 26.65 & 0.8976   \\
      \midrule
        \multicolumn{5}{l}{Muti-Domain All-in-One Method}  & \\
    \midrule
    \rowcolor{yellow!8}
       \multicolumn{2}{c}{DATPRL-IR \ \  (Ours)} & \textbf{26.22} & \textbf{0.9336} & \textbf{27.20} & \textbf{0.9018} \\
    \bottomrule
    \end{tabular}
    }
    % \end{center}
\label{table: image desnowing}
\end{table*}

\begin{figure}[t]
  \centering
  \setlength{\abovecaptionskip}{0.08cm}
  \includegraphics[width=1.0\textwidth]{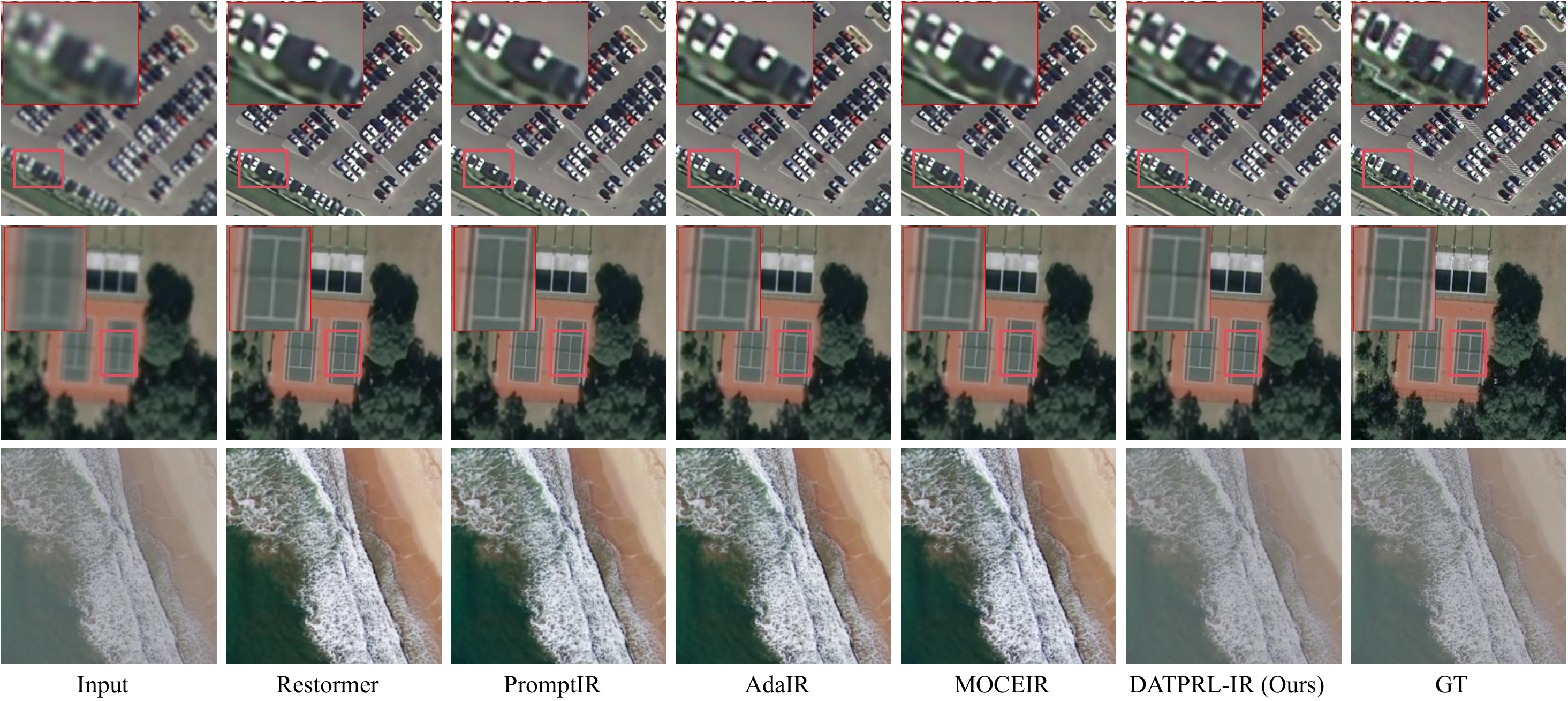}
    \caption{Generalization comparison of our DATPRL-IR with other SOTA methods on RSI deblurring task. Please zoom in for better visualization.}
  \label{fig.zeroshot_RSI_Deblur}
\end{figure}

\textbf{Generalization to unseen distributions.} We evaluate generalization performance on real-world image deraining and remote-sensing image (RSI) deblurring, both of which come from distributions different from the training data. Real-world image deraining performance is tested on SPA-Data~\citep{SPANet}, which has 1000 pairs of real-world test images. As for RSI deblurring, we randomly add varying degrees of Gaussian blur, defocus blur, and motion blur to the UCMerced~\citep{UCMerced} test set, creating 300 pairs of clean-blurred remote sensing images for testing. As shown in Table \ref{table: real-world image deraining}, our method comprehensively outperforms other AiOIR methods on both tasks. In particular, on the RSI deblurring task, it achieves an almost 2dB PSNR advantage over the SOTA methods MoCEIR, AdaIR, and PromptIR. In addition, as shown in Figure \ref{fig.zeroshot_RSI_Deblur}, the comparative methods struggle to effectively remove blur from images and have issues with task recognition errors, resulting in restored images with significant color shifts. In contrast, our method can effectively remove blur while preserving colors and other image content.

\begin{table*}
    \centering
    \Large
    % \captionsetup{skip=5pt}
    \caption{Quantitative generalization comparison between our method and other SOTA methods on real-world image deraining and RSI deblurring. The best metrics are highlighted in \textbf{bold}.}  
    \resizebox{0.58\textwidth}{!}{
    \centering
    \begin{tabular}{cc|cccc}
    \toprule
    \multicolumn{2}{c|}{\multirow{2}{*}{Task \& Dataset}} & \multicolumn{2}{c}{Real-World Deraining} & \multicolumn{2}{c}{RSI Deblurring}  \\
      & & \multicolumn{2}{c}{on SPA-Data} & \multicolumn{2}{c}{on UCMerced} \\
    \cmidrule(lr){1-6}
      Method & Year & PSNR $\uparrow$ & SSIM $\uparrow$ & PSNR $\uparrow$ & SSIM $\uparrow$   \\
    \midrule
    \multicolumn{2}{c}{Single-Task Method}  & \\
    \midrule
    MPRNet& CVPR2021 & 31.99 & 0.9263 & 26.70 &  0.7741\\
    Restormer& CVPR2022 & \textbf{32.35} & \textbf{0.9282} & 27.21 &  0.7928\\
      NAFNet & ECCV2022  & 31.99 & 0.9249 & 26.92 & 0.7864\\
      \midrule
    \multicolumn{2}{c}{All-in-One Method}  & \\
    \midrule
     Transweather & CVPR2022 & 32.21 & 0.9304 & 26.60 & 0.7684 \\

      PromptIR & NeurIPS2023  & 32.09 & 0.9254 & 26.77 & 0.7828\\

     AdaIR & ICLR2025 & 32.28 &  0.9278 & 27.22 & 0.7951\\

      MoCEIR & CVPR2025 & 32.20 & 0.9261 & 27.35 &  0.7956 \\
      \midrule
        \multicolumn{5}{l}{Muti-Domain All-in-One Method}  & \\
    \midrule
    \rowcolor{yellow!8}
       \multicolumn{2}{c}{DATPRL-IR \ \  (Ours)} &32.32 & 0.9265 & \textbf{28.91} & \textbf{0.8108} \\
    \bottomrule
    \end{tabular}
    }
    % \end{center}
\label{table: real-world image deraining}
\end{table*}

The above experiments demonstrate the two learned prompt pools capture reusable domain and task priors rather than dataset-specific patterns. When tested on unseen domains, tasks, and distributions, our model can adapt by selecting and composing prompts that best match the new data distribution, demonstrating clear transferability. This further proves the importance of modeling both shared knowledge and unique knowledge across different tasks/domains for image restoration.

\subsection{Perceptual quality comparison}
For more comprehensive evaluation, as shown in Table \ref{table: Perceptual quality comparison}, we test five perceptual metrics—CLIPIQA~\citep{CLIPIQA}, MANIQA~\citep{MANIQA}, MUSIQ~\citep{MUSIQ}, NIQE~\citep{NIQE}, and FID~\citep{FID}—under the 3-domain and 9-task setting. Across all the three tasks, our DATPRL-IR consistently achieves the best scores in CLIPIQA and FID. In image deraining task, our method almost entirely outperforms other methods, achieving the best results on four metrics: CLIPIQA, MANIQA, and MUSIQ, and FID. Overall, our method achieves better or comparable perceptual scores compared to existing methods. It worth noting that to ensure fair and meaningful evaluation, we report these metrics only on natural-image tasks, because CLIPIQA, MANIQA, and MUSIQ, etc. are pre-trained exclusively on natural-image datasets, and applying them to medical or remote-sensing images would not provide reliable or interpretable scores. Moreover, medical and remote-sensing communities~\citep{all-in-one-medir2025, Restore-rwkv-allinone-medical, amir, remote_sensing_IR1, remote_sensing_IR2} as well as AiOIR community~\citep{promptir, moceir, adair, tian, airnet} primarily focus more on fidelity-based (pixel-level) metrics due to their optimization objective and safety requirements.

\begin{table*}
    \centering
    \Large
    \caption{Perceptual quality comparison between our method and other SOTA methods on 3 domains \& 9 tasks experimental setting. The best metrics are highlighted in \textbf{bold}}.
     \resizebox{\textwidth}{!}{
    \centering
    \begin{tabular}{cc|ccccc|ccccc|ccccc}
    \toprule
     \multicolumn{2}{c|}{\multirow{1}{*}{Image Domain}} & \multicolumn{15}{c}{Natural Image} \\
    \midrule
     \multicolumn{2}{c|}{\multirow{2}{*}{\makecell{Task \& \\ Dataset}}} & \multicolumn{5}{c|}{SR} & \multicolumn{5}{c|}{Deraining}& \multicolumn{5}{c}{Deblurring}  \\
     \multicolumn{2}{c|}{\multirow{1}{*}{Dataset}} & 
       \multicolumn{5}{c|}{on DIV2K-Val} & \multicolumn{5}{c|}{on Rain100L} & \multicolumn{5}{c}{on GoPro} \\
    \cmidrule(lr){1-17}
      \multicolumn{1}{c}{\multirow{1}{*}{Method}} &\multicolumn{1}{c}{\multirow{1}{*}{Year}} & CLIPIQA $\uparrow$ & MANIQA $\uparrow$ & MUSIQ $\uparrow$ & NIQE $\downarrow$ & FID $\downarrow$  & CLIPIQA $\uparrow$ & MANIQA $\uparrow$ & MUSIQ $\uparrow$ & NIQE $\downarrow$ & FID $\downarrow$ & CLIPIQA $\uparrow$ & MANIQA $\uparrow$ & MUSIQ $\uparrow$ & NIQE $\downarrow$ & FID $\downarrow$  \\
    \midrule
     \multicolumn{2}{c}{Single-Task Method}  & \\
    \midrule
    MPRNet& CVPR2021 & 0.4561 & 0.4998 & 55.35 & 5.9716 & 104.76 & 0.7480 & 0.6967 & 70.61  & \textbf{3.1747} & 9.0289 & 0.3218 & 0.4116 & 34.01 & 5.6452 & 20.73\\
     SwinIR & ICCVW2021 & 0.4512 & 0.5059 & 58.06 & 5.9690 & 102.31 & 0.7560 & 0.6987 &  70.63 & 3.1858 & 8.3961 & 0.3142 & 0.4068 & 35.91 & 5.6113 &  25.26\\
    Restormer& CVPR2022 & 0.4634 & 0.5138 & \textbf{58.28} & 6.0288 & 102.01 & 0.7543 & 0.7010 & 70.85 & 3.2066 & 6.5893 & 0.3468 &  0.4429 & 36.96 & 5.5778 & 17.05\\
    NAFNet  &  ECCV2022  & 0.4607 &  0.5041 & 56.10 & \textbf{5.9210} & 102.45 & 0.7530 & 0.6976 & 70.68 & 3.2944 &  8.9292 & 0.3410 & 0.4410 & 37.24 & 5.4487 & 16.77\\
    \midrule
        \multicolumn{2}{c}{All-in-One Method}  & \\
    \midrule
     Transweather & CVPR2022 & 0.4500 &  0.4828 & 51.85 & 6.6003 & 102.20 & 0.6803 & 0.6682 &69.07 & 3.4578 & 33.030 &   0.3023 & 0.3472 &  27.81 & 6.1708 & 36.66\\
      PromptIR & NeurIPS2023  & 0.4585 & 0.5093 & 56.48 & 6.2211 &  101.95 & 0.7545 & 0.6998 & 70.81 &  3.2141 & 7.5467 & 0.3385 & 0.4382 & 36.39 & 5.6221 & 18.25\\
      % AMIR  & \\
       AdaIR& ICLR2025 & 0.4624 & \textbf{0.5161} & 58.18 & 6.0680 & 102.18 & 0.7556 & 0.7010 & 70.86 &  3.2149 & 6.8281 & 0.3201 & 0.4547 & \textbf{38.07} & 5.5143 & 16.79\\
        MoCEIR& CVPR2025 & 0.4604 & 0.5120 & 55.47 & 6.0852 & 102.98 & 0.7573 & 0.7011 & 70.85 & 3.2167 & 7.1322 & 0.3270 & \textbf{0.4597} & 37.87 & 5.5175 & 16.88\\
    \midrule
    \multicolumn{5}{l}{Muti-Domain All-in-One Method}\\
    \midrule
     \rowcolor{blue!8}
     \multicolumn{2}{c}{DATPRL-IR \ \  (Ours)} & \textbf{0.5558} & 0.5156 & 56.29 &  6.2490 &\textbf{101.93} & \textbf{0.7587}& \textbf{0.7033} &\textbf{70.94} & 3.2410 &  \textbf{5.1064} & \textbf{0.3485} & 0.4461 &  35.93 & \textbf{5.4463} &  \textbf{16.73}\\
    \bottomrule
    \end{tabular}
    }
\label{table: Perceptual quality comparison}
\end{table*}

\section{More Ablation Studies}

\textbf{Effectiveness on different fusion strategies.}
To evaluate the effectiveness on different fusion strategies, except for our fusion strategy where the task prompt representation is combined with the domain prompt representation and then further fused with the image features in the decoder layers, we implement another reasonable fusion strategy, domain→image→task sequential fusion variant, for comparison. We use cross-attention mechanisms to sequentially fuse domain prompt representation with image features, and then fuse the combined features with task prompt representation. As shown in Table \ref{table: Effectiveness on different fusion strategies}, although sequential fusion also achieved considerable performance, its performance still declined compared to our direct fusion. In MD-AiOIR, both domain prompt representation and task prompt representation are equally important. Our fusion method aims to learn a prompt representation that is both task-aware and domain-aware to guide the restoration network. In addition, compared to sequential fusion, our fusion method can also use the proposed adaptive gated fusion (AGF) to control the contribution ratio between backbone features and domain-aware task prompt representation, which is also very important.

\begin{table*}
    \centering
    \Large
    % \captionsetup{skip=5pt}
    \caption{Effectiveness on different fusion strategies on 3 domains \& 9 tasks experimental setting. The best metrics are highlighted in \textbf{bold}.}
    \resizebox{\textwidth}{!}{
    \centering
    \begin{tabular}{c|cccccccccccc}
    \toprule
    \cmidrule(lr){1-13}
     \multicolumn{1}{c|}{\multirow{2}{*}{Task \& Dataset}} & \multicolumn{2}{c}{Deraining}  & \multicolumn{2}{c}{Deblurring} & \multicolumn{2}{c}{CT Denoising} & \multicolumn{2}{c}{RSI SR} & \multicolumn{2}{c}{Cloud Removal} & \multicolumn{2}{c}{RSI Dehazing} \\
   & \multicolumn{2}{c}{on Rain100L}  & \multicolumn{2}{c}{on GoPro}  & \multicolumn{2}{c}{on AAPM} & \multicolumn{2}{c}{on UCMerced} & \multicolumn{2}{c}{on CUHK CR1} & \multicolumn{2}{c}{on RICE1}  \\
    \cmidrule(lr){1-13}
    % \cmidrule(lr){3-8}
      \multicolumn{1}{c}{\multirow{1}{*}{Method}}  & PSNR $\uparrow$ & SSIM $\uparrow$ & PSNR $\uparrow$ & SSIM $\uparrow$& PSNR $\uparrow$& SSIM $\uparrow$& PSNR $\uparrow$ & SSIM $\uparrow$ & PSNR $\uparrow$& SSIM $\uparrow$& PSNR $\uparrow$ & SSIM $\uparrow$\\

    \midrule
     Sequential Fusion (domain$\rightarrow$image$\rightarrow$task)   & 39.60 & 0.9865 & 29.30 & 0.8832	& 33.76 & 0.9271 & 28.27 & 0.7902 & 25.82 & 0.7577 & 26.76 & 0.9376\\
      \rowcolor{blue!8}
      Ours & \textbf{39.67} & \textbf{0.9867} & \textbf{29.57} & \textbf{0.8881} & \textbf{33.77} & \textbf{0.9273} & \textbf{28.31} & \textbf{0.7913} &  \textbf{26.00} & \textbf{0.7592} & \textbf{26.94} & \textbf{0.9347} \\
    \bottomrule
    \end{tabular}
    }
    % \end{center}
\label{table: Effectiveness on different fusion strategies}
\end{table*}

\end{document}